\documentclass[sigconf]{acmart}

\AtBeginDocument{%
  }

\usepackage{url}
\usepackage{graphicx}
\usepackage{longtable}
\usepackage{multirow}
\copyrightyear{2025}
\acmYear{2025}
\setcopyright{cc}
\setcctype{by-nc}
\acmConference[CHI '25]{CHI Conference on Human Factors in Computing Systems}{April 26-May 1, 2025}{Yokohama, Japan}
\acmBooktitle{CHI Conference on Human Factors in Computing Systems (CHI '25), April 26-May 1, 2025, Yokohama, Japan}\acmDOI{10.1145/3706598.3713440}
\acmISBN{979-8-4007-1394-1/25/04}

\begin{document}

\title{corobos: A Design for Mobile Robots Enabling Cooperative Transitions between Table and Wall Surfaces}

\author{Changyo Han}
\email{hanc@nae-lab.org}
\orcid{0000-0002-9925-3010}
\affiliation
{
  \institution{The University of Tokyo}
  \city{Tokyo}
  \country{Japan}
}

\author{Yousuke Nakagawa}
\email{nakagawa@nae-lab.org}
\orcid{0009-0003-6919-8828}
\affiliation
{
  \institution{The University of Tokyo}
  \city{Tokyo}
  \country{Japan}
}

\author{Takeshi Naemura}
\email{naemura@nae-lab.org}
\orcid{0000-0002-6653-000X}
\affiliation
{
  \institution{The University of Tokyo}
  \city{Tokyo}
  \country{Japan}
}

\renewcommand{\shortauthors}{Han, Nakagawa and Naemura}

\begin{abstract}
Swarm User Interfaces allow dynamic arrangement of user environments through the use of multiple mobile robots, but their operational range is typically confined to a single plane due to constraints imposed by their two-wheel propulsion systems.
We present \textit{corobos}, a proof-of-concept design that enables these robots to cooperatively transition between table (horizontal) and wall (vertical) surfaces seamlessly, without human intervention.
Each robot is equipped with a uniquely designed slope structure that facilitates smooth rotation when another robot pushes it toward a target surface.
Notably, this design relies solely on passive mechanical elements, eliminating the need for additional active electrical components.
We investigated the design parameters of this structure and evaluated its transition success rate through experiments.
Furthermore, we demonstrate various application examples to showcase the potential of \textit{corobos} in enhancing user environments.
\end{abstract}


\begin{teaserfigure}
  \includegraphics[width=\textwidth]{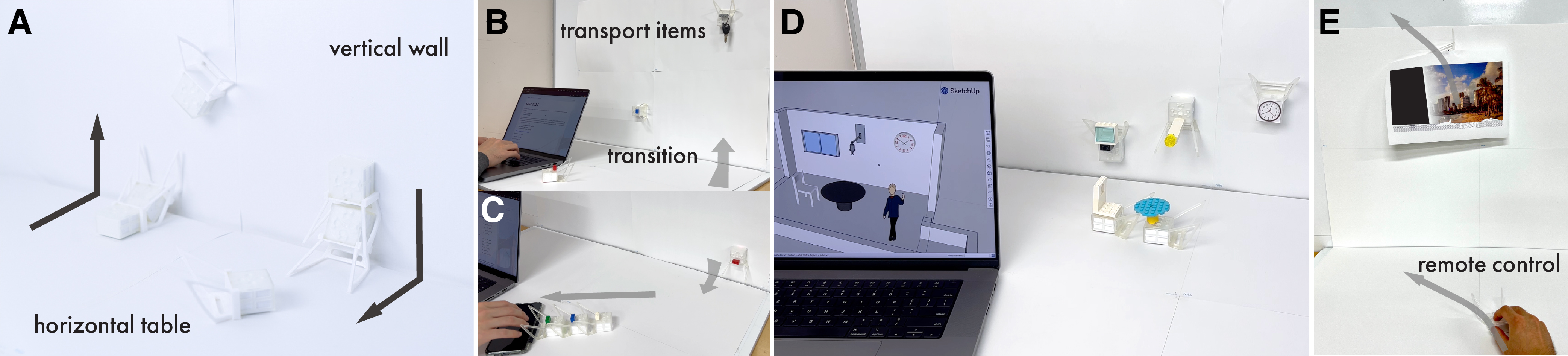}
    \caption{Overview of \textit{corobos}.
    (A) In \textit{corobos}, robots can cooperatively transition between horizontal (table) and vertical (wall) surfaces by pushing against the border.
    They can (B) move unused items to the wall to organize the workspace, (C) reallocate robots from the reserved surface to transport heavy objects, (D) physically emulate room layouts, and (E) dynamically arrange items on the wall.}
    \Description{(A) Several robots are cooperatively transitioning between table and wall surfaces. (B) A robot is carrying a key handed over by the user. (C) Three robots push a heavy smartphone with an aggregate force. (D) On the left, the laptop screen shows the layout of a room in CAD software; on the right, robots are replicating the room layout physically. (E) A robot is equipped with a poster and attached to the wall. The user is controlling the robot on the wall with another robot on the table.}
  \label{fig:teaser}
\end{teaserfigure}


\maketitle

\section{Introduction}

A diverse range of robots is being incorporated into our daily living environments, moving beyond industrial applications.
With the increasing prevalence of floor-cleaning and service robots, mobile robots will soon permeate every aspect of our living spaces, providing support for daily tasks.
Additionally, as electronic components and robots become more affordable, they are expected to shrink in size, leading to the widespread integration of numerous robots into human environments.

In this context, within the field of Human-Computer Interaction, Swarm User Interfaces (Swarm UIs), which employ multiple small robots as interfaces, have garnered considerable interest, as exemplified by the work of \textit{Zooids}~\cite{Zooids2016}, \textit{Reactile}~\cite{Reactile2018}, and \textit{ShapeBots}~\cite{ShapeBots2019}.
A robot swarm consists of numerous small robots, each with simple functionalities, that collaborate like animal swarms in nature to accomplish various tasks.
Swarm UIs interact with humans individually and convey information through the shape of the entire group, primarily driven by the robots’ movement.
For example, a group of robots can gather around a user’s phone to indicate an important call~\cite{UbiSwarm2017}.
Thus, enhancing the locomotion capabilities of swarm robots is crucial.

While the operational range of swarm robots can include vertical wall surfaces~\cite{UbiSwarm2017}, aerial spaces~\cite{Chung2018}, and clothing~\cite{Rovables2016}, their primary design is intended for horizontal surfaces, particularly tabletops~\cite{Zooids2016,HERMITS2020,disappearables2022,HapticBots2021,SwarmHaptics2019,PhysicaDIS2023}.
Despite the potential of wall surfaces for Human-Robot Interactions~\cite{Yan2021,UbiSwarm2017}, these robots typically require manual relocation by humans or additional infrastructure (e.g., ramps).
This limitation is mainly due to the basic locomotion mechanisms of most mobile robots, particularly two-wheel propulsion, which restricts self-transition between surfaces.
To our knowledge, no mobile robots designed for Swarm UIs currently have the capability to autonomously transition between tabletops and walls.

Unlocking the ability for swarm robots to autonomously transition between horizontal and vertical surfaces—like tables and walls—would expand their capabilities, enabling spatially richer interactions.
For instance, objects could be stored or displayed on walls, freeing table space, or additional robots could be summoned from the wall to the table for cooperative tasks.
Moreover, robots could seamlessly move from a tabletop to a wall and then to another tabletop to enrich communication between users within the same workspace.

In this paper, we extend our previous work~\cite{corobos_UIST} and introduce \textbf{\textit{corobos}}, a novel proof-of-concept design for mobile robots that can transition between horizontal tabletops (where hands-on work and user interactions take place) and vertical walls (which are harder to reach but well-suited for displays, storage, and shared visualization).

Building on the concept of Tangible Bits~\cite{Ishii1997} and Embodied Interaction \cite{Dourish2001_where}, \textit{corobos} leverages the physical environment to shape user attention.
By enabling robots to transition between a table and a wall, we integrate digital objects into the user’s spatial and bodily routines, influencing how attention is allocated.
The table, within easy reach, invites hands-on manipulation, while the wall positions objects at a more distant location, prompting users to shift their gaze and stance.

Taking inspiration from the cooperative behavior observed in nature, such as army ants that bridge large gaps by forming interlinked bridges~\cite{ArmyAntBridge2015}, our robots leverage collaborative behavior of the robots to transition across surfaces.
We developed a passive 3D-printable attachment that can be easily fabricated and is compatible with commercially available two-wheeled mobile robots.
This passive mechanical design eliminates the need for additional powered components, such as motors, simplifying its integration into Swarm UIs.
With this attachment, one robot can transition from one surface to another, aided by a second robot pushing from behind.
 This method leverages the robots’ inherent capabilities, minimizing alterations to wall surfaces or desktop environments.

The contributions of this paper include:
\begin{itemize}
    \item A design concept and mechanism that enabling two-wheeled swarm robots to transition between horizontal and vertical surfaces.
    \item A systematic and empirical validation of the design parameters of the attachment and implementation using an existing robotic toy platform.
    \item A series of application examples demonstrating the table-wall transition with enhanced interaction capabilities.
\end{itemize}

\section{Related Work}

\subsection{Swarm User Interfaces}

Distributed and closely cooperating groups are commonly referred to as swarms, which can accomplish tasks beyond the capabilities of individual members through collaboration.
Swarm robotics is an approach inspired by the collective behavior of animal groups, aiming to design robust, flexible, and scalable group behaviors by coordinating a large number of robots using simple rules and local interactions~\cite{Brambilla2013}.
The number of swarm robots varies depending on the research context; for instance, \textit{Kilobot}~\cite{Kilobot2014} presents a low-cost, easy-to-assemble swarm robot that can be used to test swarm algorithms with thousands of robots.
However, the maximum speed of \textit{Kilobot} is only 1 $cm/s$, which is relatively slow and unsuitable for real-time interaction with users.
The size of swarm robots ranges from very small, approximately 4 $mm$ in size~\cite{Wu2022}, to those similar in size to cleaning robots~\cite{RoomShift2020}.

In the field of Human-Computer Interaction, research involving swarm robots has been ongoing since the late 2010s.
Swarm UIs are a relatively new category of interfaces in which multiple self-propelled robots respond to user input and environmental changes~\cite{Zooids2016,Suzuki2022,HERMITS2020,UbiSwarm2017,ShapeBots2019,disappearables2022,HapticBots2021}.
One advantage of Swarm UIs is their composition of numerous interchangeable elements, enabling flexible adjustments to the overall size and shape of the interface.
For example, \textit{Zooids}~\cite{Zooids2016} can dynamically change the overall shape of the swarm by moving collectively, providing information through the spatial arrangement of robots.
Interactions in which individual robots control the behavior of the entire group are also possible.

Due to these characteristics, Swarm UIs are being explored for various applications, such as actuated tangible user interfaces~\cite{HERMITS2020,disappearables2022}, providing haptic feedback~\cite{SwarmHaptics2019,HapticBots2021}, creating physical displays~\cite{Alonso-Mora2011,Alonso-Mora2012,Alonso-Mora2015}, embodying the human body parts~\cite{SwarmBody}, manipulating objects on tabletop surfaces~\cite{Push-That-There2024}, and assisting digital fabrication machines~\cite{FabRobotics2024}.

Many Swarm UIs are designed primarily for tabletop use~\cite{Reactile2018,ShapeBots2019,Hiraki2016,disappearables2022,HERMITS2020}, as tabletops are considered suitable locations for human interaction.
For example, \textit{HERMITS}~\cite{HERMITS2020} expands the capabilities of individual robots by connecting detachable external parts, called mechanical shells, to modified small robots, thereby demonstrating the application of swarm robots through cooperative work.

Building on previous work in Swarm UIs, \textit{corobos} aims to extend the operational range of these interfaces by introducing a mechanism that facilitates transitions between tabletop and wall surfaces.

\subsection{Robots Adhering to Surfaces}
Swarm UIs primarily utilize two-wheeled robots on tabletops, but there are also robots with alternative locomotion capabilities suitable for different environments.

For instance, \textit{Griddrones}~\cite{Griddrones2018} are Swarm UIs that operate in the air.
These cube-shaped drones can create 3D shapes as voxels, with their size limited only by the available indoor space.
However, drones consume significant power for flight, making it challenging for them to remain airborne for extended periods.

Robots that adhere to walls or ceilings have been extensively investigated in the field of robotics~\cite{Inoue2006,Ahmed2022,Panich2010,Tripillar2011}.
Methods for adhering to walls and ceilings can be broadly classified into four categories: vacuum suction, adhesive materials, mechanical mechanisms, and magnetic adhesion.

Vacuum suction~\cite{Panich2010,suctioncup2010,EpidermalRobots} or air thrust~\cite{Vertigo2015} generates an attractive force through air pressure.
While this method can adhere to various flat wall materials, it has disadvantages, such as noise and increased device size, making it unsuitable for indoor swarm robots.

The adhesive material approach~\cite{Waalbot2007,Kim2008,Yan2021,HawkesGeckoHuman2015} uses dry or wet sticky substances to adhere to walls.
Although generally quieter than vacuum suction, its adhesion force is weaker because it depends on the surface friction of the wall material.

Mechanical mechanisms~\cite{Inoue2006,CLASH2011} allow robots to adhere by inserting claw-shaped parts into the unevenness of the wall surface, offering strong adhesion force and high reliability of movement.
However, this approach often results in larger, multi-legged robots due to its complexity.

Magnetic adhesion~\cite{iRobot2023,UbiSwarm2017,Ahmed2022,Takada2017,Matsumura2019,AeroRigUI_CHI2023,ThrowIO_CHI2023,Tripillar2011,R-Track2021,ThreadingSpace2024} is a commonly used method in various applications due to its silent operation and reliability on ferromagnetic surfaces.
For industrial purposes, inspection robots for steel structures (such as bridges) use magnetic adhesion to adhere to various surfaces~\cite{Ahmed2022}.
In educational settings, the iRobot Root\texttrademark~\cite{iRobot2023} adheres to whiteboards using magnets on its underside, allowing it to perform tasks like drawing and erasing lines with an attached pen.

In examples like \textit{Rovables}~\cite{Rovables2016} and \textit{Calico}~\cite{Calico}, robots can move on clothing.
These robots climb vertically without modifying the clothing by clamping their wheels onto the fabric using magnets or specially designed rails.

Among Swarm UIs that move on walls, \textit{UbiSwarm}~\cite{UbiSwarm2017} is a notable example.
In \textit{UbiSwarm}, magnets attached to the robot’s bottom surface allow them to adhere to ferromagnetic wall surfaces.

\textit{AeroRigUI}~\cite{AeroRigUI_CHI2023}, \textit{ThrowIO}~\cite{ThrowIO_CHI2023}, and Threading Space~\cite{ThreadingSpace2024} also use magnetic adhesion to attach objects and robots to ceilings, enabling 3D mid-air interactions, throwing and catching interactions, or altering spatial perception with physical threads.

Although installing strong ferromagnetic materials like whiteboards (steel plates) on the wall surface is necessary, magnetic adhesion allows for smaller size, passive (non-powered) adhesion, noise-free operation, and stronger adhesion force than adhesive materials.
Consequently, for small-sized mobile robots in Swarm UIs, magnetic adhesion is a suitable choice, and thus, we have adopted this method for \textit{corobos}.

\subsection{Surface-Transitioning Robots}
To transition between perpendicular surfaces, such as a tabletop and a wall, a robot must rotate 90 degrees while adhering to the surface, necessitating a unique mechanism.
For example, Ahmed et al. developed a robot~\cite{Ahmed2022} for inspecting structures made of ferromagnetic materials.
This robot is equipped with joints that connect leg-like parts with magnets, enabling it to adhere to and transition on structures with noncontinuous surfaces.
Another approach involves using magnetic wheels~\cite{Matsumura2019,Tripillar2011}, which allow a robot to adhere to a structure’s surface and move freely in any direction by rotating the wheels.
However, these methods result in complex, high-performance robots, making them unsuitable for Swarm UIs.

Conversely, \textit{FreeBOT}~\cite{Liang2020} consists of multiple spherical robots that adhere to each other using magnets, allowing their overall shape to change dynamically.
\textit{FreeBOT} can also adhere to and transition on ferromagnetic walls while rotating on its own and can even ride on top of another robot.
However, this transition method is not ideal for interactions involving humans or objects, such as pushing or pulling in one direction, and accurately estimating the position is challenging due to the robots’ form factor.

In \textit{UbiSwarm}~\cite{UbiSwarm2017}, researchers proposed the concept of Ubiquitous Robotic Interfaces (URIs) to make robots ubiquitous and capable of interacting with humans and their environment.
Their vision involves groups of robots seamlessly transitioning from walls to tabletops and between different rooms at a speed comparable to the human eye’s refresh rate, providing virtually unlimited interaction space.
Although they mentioned a method involving a magnetic slope between walls and tabletops as a candidate for transitioning, it has not yet been realized in the paper.

\textit{R-Track}~\cite{R-Track2021} features modular robots equipped with magnetic tracks that can transition from wall to wall by connecting three robots in sequence.
These robots are capable of navigating both internal and external angles between two perpendicular surfaces.
However, their design is focused on inspection tasks, with a size of about 150 mm, making them less suitable for Swarm UI applications, which typically require smaller, more compact robots.
Additionally, \textit{R-Track} incorporates dedicated mechanisms like external connectors to facilitate wall-to-wall transitions, which, while effective, are unnecessary for simpler table-to-wall transitions in Swarm UI scenarios.
The simpler structure of \textit{corobos} provides a more practical approach for these applications by minimizing complexity and focusing on specific user interaction needs.

In \textit{(Dis)Appearables}~\cite{disappearables2022}, a small two-wheeled robot can transition between horizontal surfaces of different heights by using a ferromagnetic material ramp installed between the two surfaces.
While this method can be applied to \textit{corobos} system, it restricts transition opportunities to locations with ramps, and the ramp’s continuous presence on the desktop may interfere with human activities.

In our approach, we aim to develop a transition method that requires minimal modification to both the robots and the desktop environment.

\section{Design Concepts of corobos}
In this paper, we introduce a prototype design for swarm robots, \textit{corobos}, which enables cooperative transitions between table and wall surfaces.
The concept of cooperative operation for these robots is derived from the principle of \textit{self-assembly} in swarm robotics by Gross et al.\cite{AutonomousSelfAssemblySwarm2006}.
While such autonomous reconfigurability is well demonstrated in quadruped robots by Ozkan-Aydin et al.\cite{Yasemin2021}, it is challenging to apply it to Swarm UIs due to the complex mechanisms involved.

As previously mentioned, robots for Swarm UIs are often limited to tabletop use, and one potential approach for small two-wheeled mobile robots to transition between a tabletop and a wall involves installing a ferromagnetic ramp between the two surfaces, allowing the robots to adhere while transitioning.
However, this method restricts transitions to specific locations with ramps and presents a drawback by continuously obstructing human activities when the ramp is on the table.
Ideally, robots should be able to transition using their inherent capabilities without significant modifications to the wall or tabletop environments.

Instead, we propose a concept where multiple robots cooperate to achieve seamless transitions.
To accomplish this, \textit{corobos} uses additional attachments on existing two-wheeled mobile robots.
Through coordinated movements, these robots can transition between surfaces without relying on complex mechanisms or external structures.

\subsection{Design Implications}
According to Dourish’s framework of embodied interaction, physical spaces and the objects they contain are not passive backdrops but integral components that shape interaction \cite{Dourish2001_where}.
\textit{corobos} embodies this principle by allowing objects to occupy either the tabletop or the wall.
On the table, objects become immediate and manipulable, supporting tasks that demand focused attention and direct control.
On the wall, these same objects adopt a peripheral role, remaining visible yet less demanding, aligning with the natural way humans navigate attention through subtle bodily repositioning and spatial orientation.
This fluid movement between surfaces thus structures users’ cognitive processes, allowing them to smoothly shift from focused engagement (when objects are within arm’s reach) to peripheral awareness (when objects are placed out of direct reach yet remain in their field of view).

\subsection{Transition Mechanism}

\begin{figure}[tbhp]
\centering
\includegraphics[width=1.0\columnwidth]{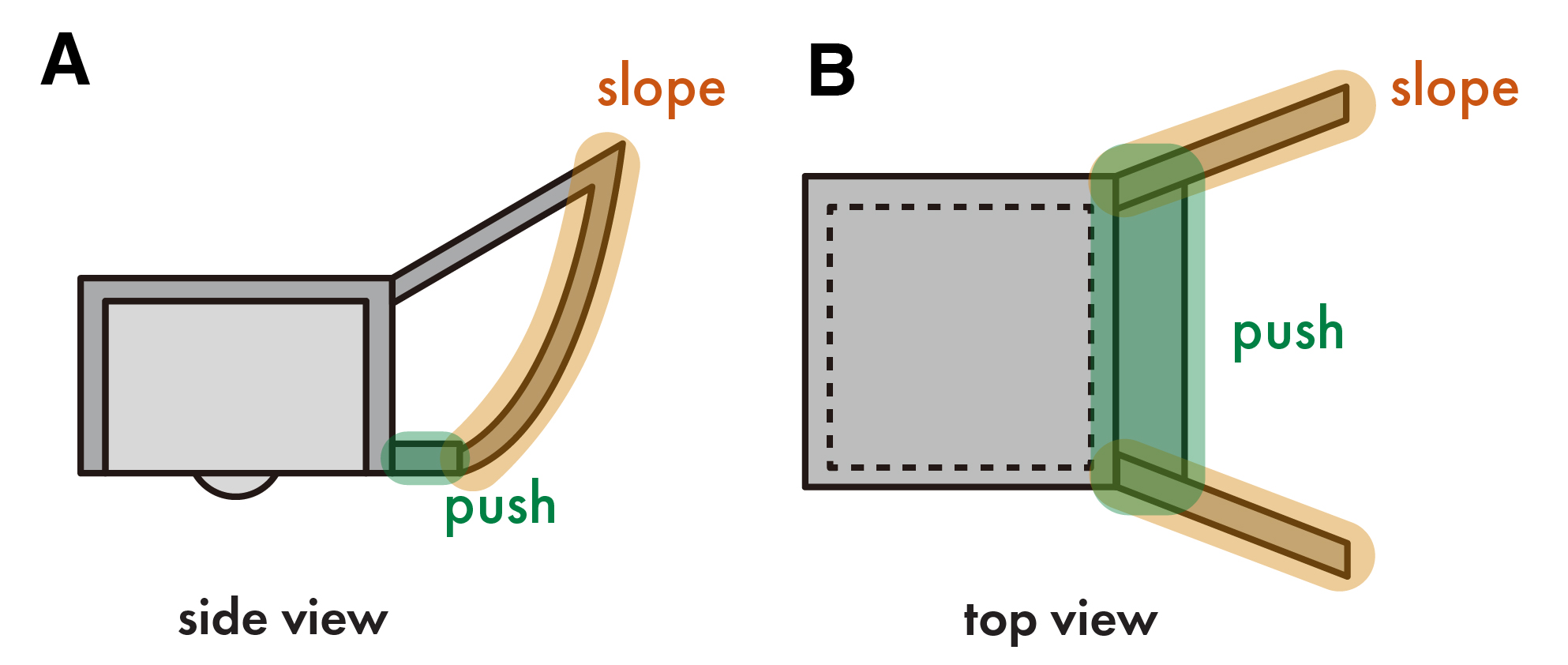}
\caption{Basic shape of the attachment. It has two functional components called \textit{slope} and \textit{push}.}
\Description{Drawing of the attachment, two functional components (push, slope) highlighted with colors.}
\label{fig:attachment_shape}
\end{figure}

In \textit{corobos}, two robots work together to achieve a 90-degree rotation by pushing the transitioning robot against the wall or table.
A basic design of the attachment is shown in Figure~\ref{fig:attachment_shape}.
The attachment consists of two functional components: \textit{push} and \textit{slope}.
The push component is responsible for pushing the lower rear part of the transitioning robot, while the slope component facilitates a 90-degree rotation when pushed.
The push component is positioned at the front of the robot, in front of the slope component.

\begin{figure}[tbhp]
\centering
\includegraphics[width=1.0\columnwidth]{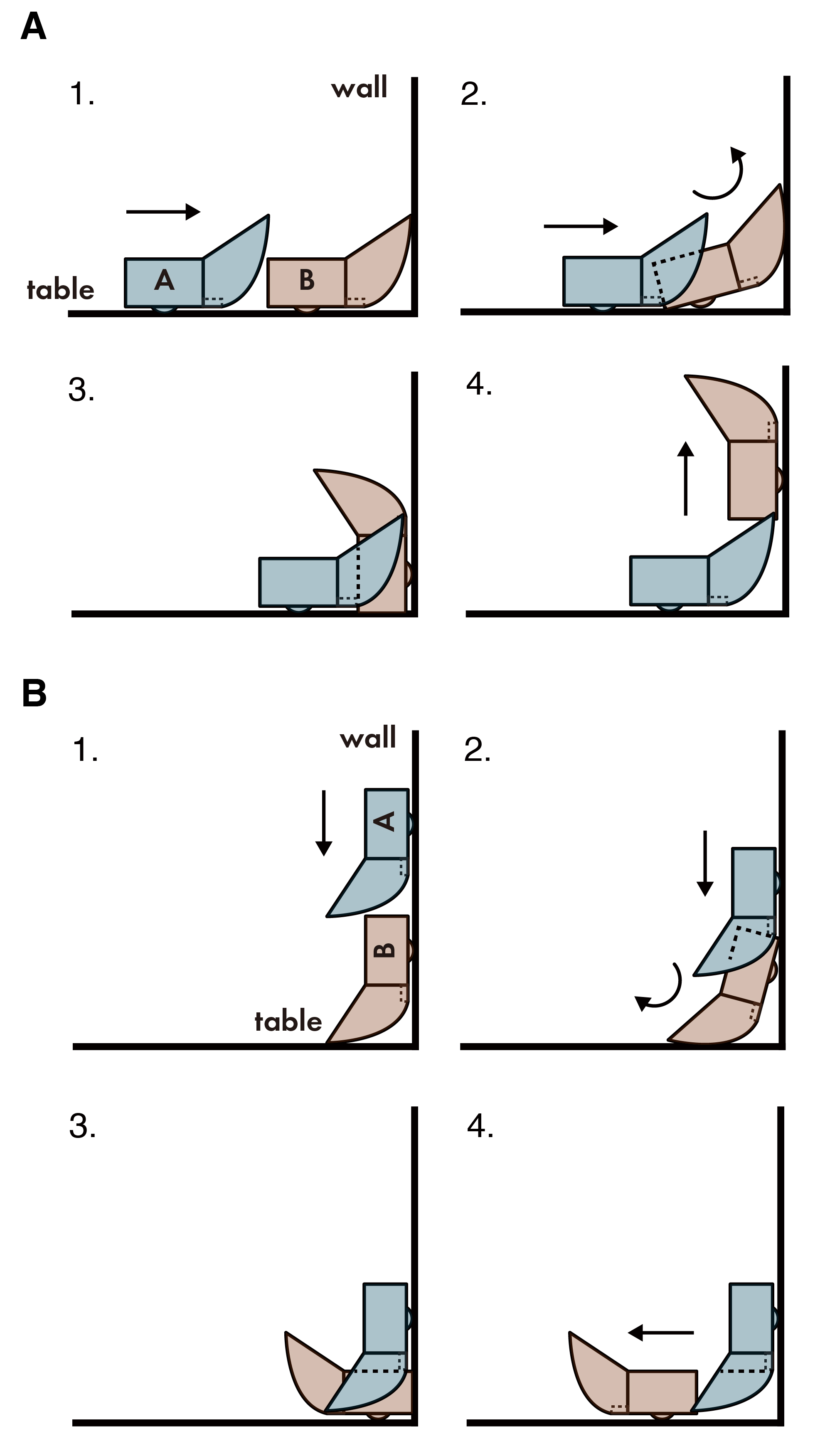}
\caption{Transitioning process of \textit{corobos}:
(A) Table-to-wall transition, (B) Wall-to-table transition.}
\Description{(A) Description of the ascending transitioning process. As the helper robot in the rear pushes the transitioning robot in the front, the transitioning robot rotates 90 degrees and attaches to the wall. (B) Description of the descending transitioning process, similar to (A) but from the wall to table.}
\label{fig:transition_process}
\end{figure}

For the transition from the tabletop to the wall, the movement occurs as illustrated in Figure~\ref{fig:transition_process}~(A).
In the following explanation, the helper robot (blue) is referred to as A, and the transitioning robot (red) is referred to as B.

\begin{enumerate}
\item Position robots A and B in a straight line on the tabletop near the wall where the transition will occur.
\item Robot A pushes robot B, causing B to press against the wall and rotate 90 degrees along with the \textit{slope} component.
\item Robot B adheres to the ferromagnetic wall using the magnet attached to its bottom.
\item Once the transition is complete, robot B can move freely on the wall.
\end{enumerate}

Similarly, the transition from wall to table can be performed in the opposite direction, as shown in Figure~\ref{fig:transition_process}~(B).

Since all robots are equipped with a common attachment, they can transition between tabletops and walls using any pair of robots.
This flexibility enhances the overall system by allowing one robot to replace another, even if one fails to function.

\subsection{Design Space}

\begin{figure*}
    \centering
    \includegraphics[width=1.0\textwidth]{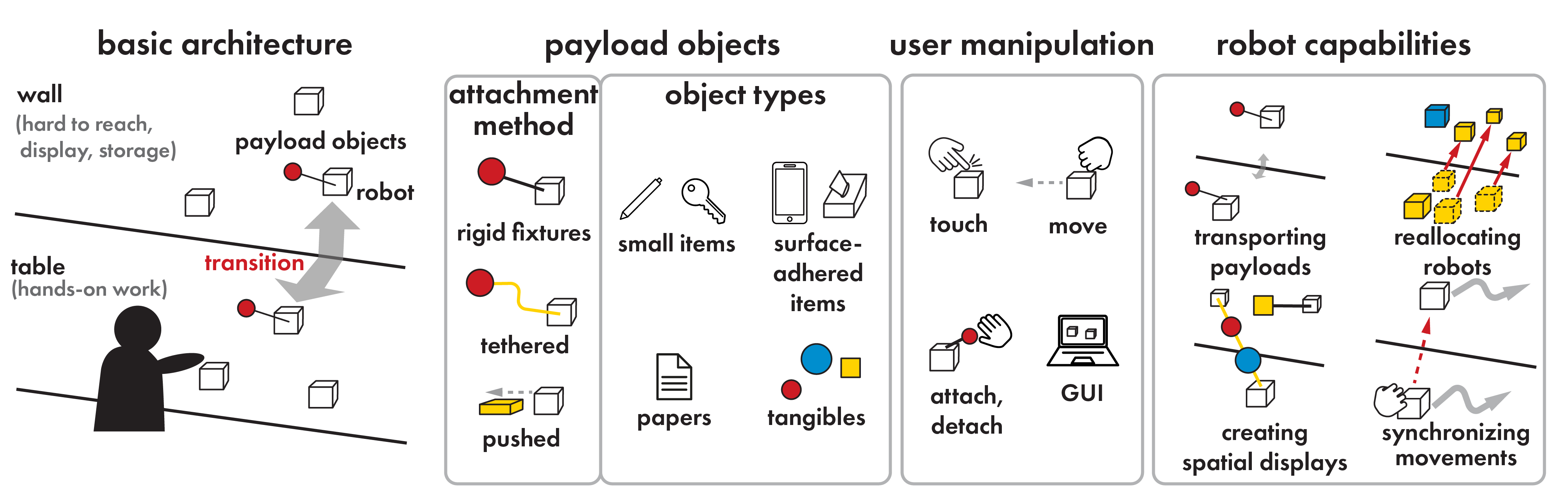}
    \caption{Design space of \textit{corobos}: surface-transitioning swarm robots.}
    \Description{Illustrations to show the design space of \textit{corobos}. It shows the basic architecture of the system, payload objects categorized in attachment and object types, user manipulation of the robots, and their capabilities.}
    \label{fig:design_space}
\end{figure*}

We introduce the design space of \textit{corobos} as shown in Figure~\ref{fig:design_space}.
Through the design space, we intend to share the new design possibilities of swarm UIs for researchers and designers.

\subsubsection{Basic Architecture}
\textit{corobos}' architecture integrates a table, a wall, and payload objects on robots to facilitate diverse interactions.

\textbf{Table:} Serving as the primary user interaction surface, the table supports conventional hands-on workspace activities alongside interactive robot manipulation.

\textbf{Wall:} The wall is adjacent to the table so the robots can make transition between them.
It also provides a more proper space for data display than the table because the wall is usually not populated with objects.
It also offers storage for robots, helping clear the workspace on the table.
User-unreachable areas (higher part of the wall) can be utilized for the robots.

\textbf{Robots:} The mobile robots are the key components of Swarm UIs.
These mobile units perform actions and navigate across surfaces, thus diversifying the interactive environment.

\textbf{Payload Objects:} The payload objects are carried by the robots to enable interaction within the user environment.

\subsubsection{Payload Objects}
Payload objects are vital elements in \textit{corobos}, with robots transporting them across or within surfaces.

\textbf{Attachment Method:}
Payload objects can be attached to the robot in various ways.
On attaching the payload objects directly to the robot, we can use rigid fixture like rod and fix the payload on the end of the rod.
The objects can also be attached to the robot with flexible interface like strings.
These configuration allows the payload object to transition between the surfaces along with the robots.
Other than directly attaching the payload object, the robots can push the objects on the table or wall to make them move within the surface.

\textbf{Object Types:}
Payload objects vary, from small items like keys which can be carried from to heavier objects requiring collective robot effort.
They may also include informational papers or tangibles for data visualization.
Small items such as keys, which can be carried from surface to surface, can be directly attached to the robots.
For surfaced-adhered items which cannot be carried between the surfaces, they can be pushed by the robots for reallocation within the existing surface.
Heavy items which a single robot cannot carry, can be moved with the cooperation of multiple robots.
The robots can also be attached with a lightweight and large-surface area items like papers.
Also, Tangibles can be attached for data physicalization or spatial display purpose.

\subsubsection{User Manipulation}

Users can manipulate and interact with the robots with many different ways.
They can touch the robot to trigger certain action, or move the robot to a certain position to input spatial and temporal information.
Users can attach or detach payload objects to the robots to make them perform certain tasks.
Also, the robots can indirectly operated by GUI.

\subsubsection{Robot Capabilities}
Outlined below are foundational actions of \textit{corobos}, with detailed applications discussed in Section~\ref{sec:applications}.

\textbf{Transporting Payload Objects:}
In \textit{corobos}, robots can transport payload objects between table and wall surfaces.
Robots can securely hold items with the equipped adaptors while transitioning between surfaces.

This capability allows \textit{corobos} to dynamically relocate objects, freeing up workspace or bringing items into easy reach as needed.
It enhances flexibility and organization, making the system ideal for environments that require frequent reorganization or adaptive use of space.

\textbf{Reallocating Robots:}
The number of robots on different surfaces can be dynamically adjusted based on task requirements or user needs.
When robots are not immediately needed on the tabletop, they can be moved to the wall to remain in standby mode until they are required again.
If a specific task requires more robots on a certain surface to create a visual display or organize items, robots can be reassigned from a surface to surface to meet this need.
This flexibility allows for efficient use of space, preventing clutter on the workspace and ensuring that robots are optimally positioned for their next tasks.

Reallocating robots between surfaces is particularly beneficial when the number of available robots is limited.
By dynamically adjusting their positions, tasks that require more robots in specific areas can be effectively managed without the need for additional robots.
This adaptability ensures that a limited number of robots can handle a wide range of tasks by being strategically repositioned as needed, maximizing their utility and minimizing idle time.

\textbf{Creating Spatial Displays:}  
\textit{corobos} enhances the creation of physical spatial displays by tethering the tabletop and wall surfaces with strings, forming lines in space~\cite{ThreadingSpace2024}.
Objects can be attached to these strings, allowing them to float in mid-air.  
Unlike conventional Swarm UIs, which can only create shapes within a 2D plane, \textit{corobos} enables the formation of complex, volumetric shapes in 3D space.

\textbf{Synchronizing Movements:}  
When a robot moves to a surface that is out of reach for users, such as a higher part of the wall, robots on accessible surfaces (like the tabletop) can be used to remotely control those on other surfaces.
This control can be established in a 1:1 relationship or extended to a 1:N relationship, depending on the interaction scenario.

\section{Implementation of corobos}  
In this section, we describe the implementation process using the off-the-shelf robotic toy platform, toio\texttrademark~\cite{toio}.  
Hardware and communication specifications of the toio robots are available for development and customization\footnote{\url{https://toio.github.io/toio-spec/en/}}.  

\subsection{Design of the Attachment}  

\begin{figure}[tbhp]  
    \centering  
    \includegraphics[width=1.0\columnwidth]{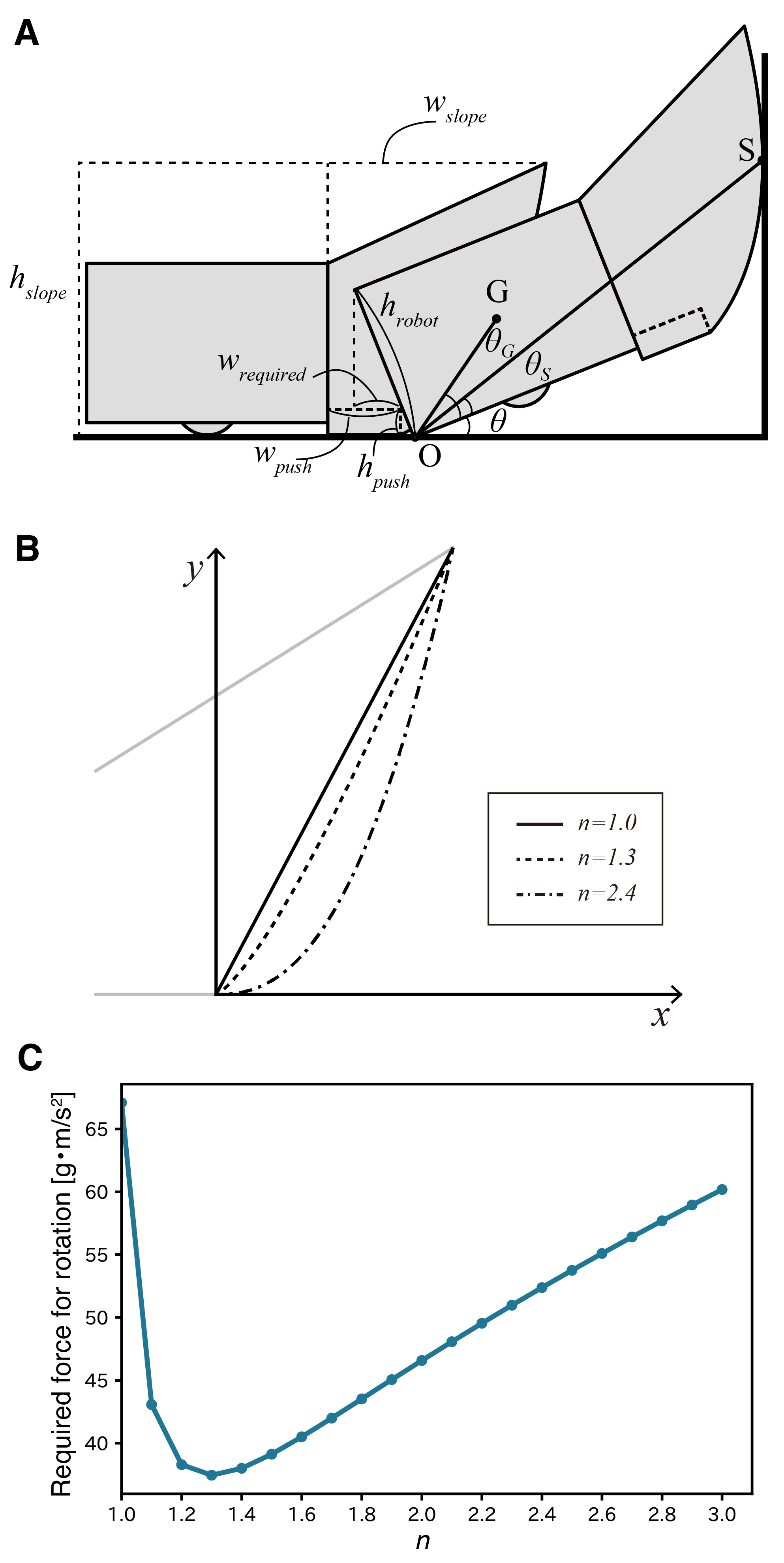}  
    \caption{  
    Parameters for determining the curvature of the slope component:  
    (A) Parameters for computing the required momentum for transitioning.  
    (B) Slope configurations corresponding to each parameter.  
    (C) Required force for rotating the transitioning robot.}  
    \Description{Illustration of the parameters for designing the slope component. (A) An illustration showing the dimensions (lengths, angles) of the transition process. (B) A picture showing the curvature for n=1.0, 1.3, 2.4 for equation (3), (C) a graph describing the required force for rotating the transitioning robot.}  
    \label{fig:design_slope}  
\end{figure}  

We describe the design process of the attachment in detail, taking the dimensions of the toio robot into account.  
Given that we use toio robots, we assumed a cube-shaped robot as the design target.  

\subsubsection{Push Component}  
Figure~\ref{fig:design_slope}~(A) shows the parameters considered in the transitioning process.  
The labels $G$, $O$, and $S$ denote critical points of interaction during the robot's transition: $G$ corresponds to the robot's center of gravity, $O$ indicates the contact point between the ground and the rear of the robot, and $S$ marks the contact point between the wall and the slope component of the attachment.  
When pushing the transitioning robot with the push component, the upper part of the push component comes into contact.  
As it is pushed, the transitioning robot gradually tilts; therefore, it is important to set the width of the push component, $w_{push}$, so that it does not come into contact with the upper surface of the transitioning robot.  
To achieve this, we define the necessary length as $w_{required}$, which is the width of the slope component.  
Assuming that the robot and push component are rectangular parallelepipeds, and defining the height of the robot as $h_{robot}$ and the rotation angle of the robot as $\theta$, we can calculate $w_{required}$ as follows:  

\begin{equation}  
w_{required} = h_{robot} \sin \theta - h_{push} \tan \theta \label{eq:w_required}  
\end{equation}  

The maximum value of $w_{required}$ from $\theta=0$° to $\theta=90$° is the minimum length required for $h_{push}$.  
This occurs when:  

\begin{equation}  
\theta = \cos^{-1} \left(\frac{h_{push}}{h_{robot}}\right)^\frac{1}{3}  
\label{eq:w_required_theta}  
\end{equation}  

By substituting $h_{push}=5.0$ mm and $h_{robot}=25.8$ mm (including the thickness of the attachment of 1.5 $mm$) into equation \ref{eq:w_required_theta}, we obtain $\theta$.  
Subsequently, substituting this value into Equation~\ref{eq:w_required}, we find $w_{required}=15.0$ $mm$.  
This is the maximum value of the required length, so we set it as $w_{push}$.  

The length of $w_{slope}$ was set equal to the height of the robot to prevent interference with the transitioning robot when pushing it to rotate 90 degrees.  
With $h_{slope}$ at the same height as the robot, there was insufficient rotational moment for the transition to occur.  
After considering the adequate height and the center of gravity of the attachment through multiple prototypes, $h_{slope}$ was set at twice the height of the toio, which is 48.6 $mm$.  

\subsubsection{Slope Component}  
Next, we explain the shape of the slope component of the attachment.  
We designed the slope component to be wider than the robot's width to prevent the transitioning robot from colliding with it during rotation.  
It was also designed to account for position misalignment due to localization errors of ±5 coordinates, where 1 coordinate of the localization system (toio Mat) corresponds to 1.42 mm, resulting in a width of approximately 62 mm, as shown in Figure~\ref{fig:attachment_cad}(B).  
Hereafter, we use xy-coordinates as viewed from the side of the attachment, as shown in Figure~\ref{fig:design_slope}(B).  
The origin is at the lower tip of the push, and the point $(w_{slope},h_{slope})$ is the vertex of the slope.  
To enable smooth rotation of the robot, a convex shape is preferable for the slope.  
Therefore, we used the curve represented by the following equation (\ref{eq:slope}), where $n$ is a real number greater than or equal to 1.0.  

\begin{equation}  
y=h_{slope} \cdot \left(\frac{x}{w_{slope}}\right)^n \;\;\;\; (0<x<w_{slope}) \label{eq:slope}  
\end{equation}  

To determine the optimal shape of the slope, we varied $n$ from 1.0 to 3.0 and calculated the minimum force required for the table-to-wall transition.  
The wall-to-table transition was not considered, as gravity assists robots in making this transition more easily.  
Specifically, we divided the x-coordinate from 0 to $w_{slope}$ into 1000 equal intervals and calculated the force required for rotation based on the moment balance when the robot in transition makes contact with the wall at each point on the curve using Equation (\ref{eq:power}), determining the maximum force required.  
Surface friction was not considered for simplicity.

\begin{equation}  
P = \frac{m g \cdot \overline{OG} \cdot \cos(\theta+\theta_G)}{\overline{OS} \cdot \sin(\theta +\theta_S)} \label{eq:power}
\end{equation}

Here, $m$ is the mass of the robot, $g$ is the gravitational acceleration, and $\overline{OG}$ and $\overline{OS}$ are the distances from the center of rotation to the center of gravity and the contact point with the wall, respectively.
Moreover, $\theta$ is the rotation angle of the robot, $\theta_S$ is the angle between $\overline{OG}$ and the robot's bottom surface, and $\theta_G$ is the angle between $\overline{OG}$ and the robot's bottom surface.
In this case, the center of gravity was calculated as the center of the toio robot.

The results are shown in Figure~\ref{fig:design_slope}(C).
From these results, we inferred that when $n=1.3$, the required rotational force is minimal, leading to a more seamless transition.

\subsection{Experiments}  
Based on the design outcomes, we fabricated the attachment and conducted comparative experiments to evaluate the transition success rate and transition time.

\subsubsection{Experimental Setup}  
The experiments were conducted using three different attachments, as shown in Figure~\ref{fig:design_slope}(B): $n=1.0$, $n=1.3$, and $n=2.4$, corresponding to the maximum, minimum, and median required forces, respectively.  
The experimental conditions were as follows:  
First, we affixed the toio Mat for developers\footnote{\url{https://toio.io/blog/detail/20200423-1.html}} onto both a wooden tabletop and a ferromagnetic wall surface (whiteboard).  
The toio Mat enables localization of toio robots using micro dot patterns and the infrared camera embedded in the robots.

We controlled the robots using a script in Unity.  
Two robots were used for the experiment: a helper robot and a transitioning robot.  
These robots were positioned side-by-side at the corner where the table meets the wall.  
During each transition, both robots were set to a motor speed of 100 (approximately 430 rpm, according to the toio specification) and directed toward the corner.  
We conducted 100 trials for each case.  
A successful transition was defined as the robot rotating 90 degrees within five seconds and recognizing the position ID of the toio Mat on either the wall or the table; otherwise, the attempt was considered a failure.  
The transition time was measured from the start of pushing until the robot recognized the target surface.  
Given that the system operates at 60 fps, a minimum timing error of approximately 0.017 seconds is expected.

\subsubsection{Results}  

\begin{table}[tbhp]  
    \caption{Relationship between attachment shape and transition success rate}  
    \label{table:attachment_success_rate}  
    \Description{A table showing the transition success rate for n=1.0, 1.3, 2.4.}  
    \centering  
    \begin{tabular}{crr}  
      \hline  
        & \multicolumn{2}{c}{Transition success rate ($\%$)}\\  
        $n$   & Table $\rightarrow$ Wall & Wall $\rightarrow$ Table\\  
      \hline \hline  
      ~~1.0 ~~  & 51  & 100 \\  
      ~~1.3 ~~  & 100   & 100 \\  
      ~~2.4 ~~  & 100  & 100 \\  
      \hline  
    \end{tabular}  
  \end{table}  

\begin{figure}[tbhp]  
    \centering  
    \includegraphics[width=1.0\columnwidth]{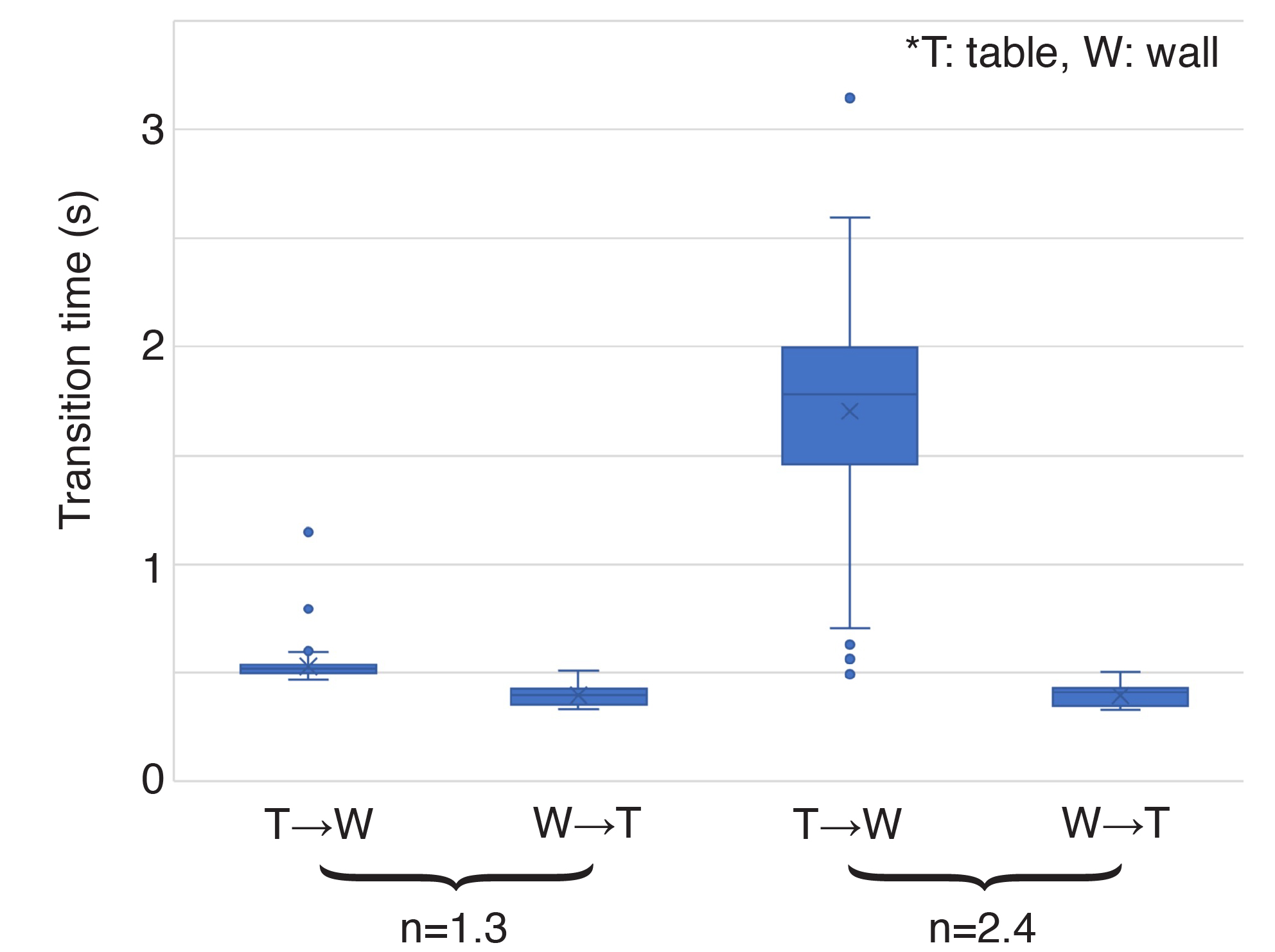}  
    \caption{  
        Experimental results of transition time.  
    }  
    \Description{A boxplot showing the transition time for n=1.3 and 2.4. For n=1.3, the transition time is consistently less than one second, while for n=2.4, the transition time from the table to the wall is over 1.5 seconds with higher variation.}  
    \label{fig:experimental_results}  
\end{figure}  

The results of the transition success rate measurements are presented in Table~\ref{table:attachment_success_rate}.  
All attachments achieved a 100 \% success rate when transitioning from the wall to the table.  
This can be attributed to the fact that once the slope component makes contact with the tabletop and initiates rotation, the magnetic adhesion weakens, and gravity aids the rotation.  
Conversely, for transitions from the tabletop to the wall, 100 successful transitions were achieved for $n=1.3$ and $n=2.4$, whereas only 51 were successful for $n=1.0$.  
This is likely due to the abrupt change in inclination at the boundary between the push and slope components.

Figure~\ref{fig:experimental_results} summarizes the results for the transition time concerning $n=1.3$ and $n=2.4$.  
The horizontal line inside the box is the median.
The box spans from the 25th to the 75th percentile, capturing the middle 50 \% of values.
The whiskers extend to approximately cover 95 \% of the data, and any dots beyond the whiskers indicate outliers.
For transitions from the wall to the tabletop, both $n=1.3$ and $n=2.4$ showed no significant difference, with an average time of 0.40 seconds and a standard deviation of 0.001.  
For transitions from the tabletop to the wall, the average time was 0.53 seconds for $n=1.3$ and 1.70 seconds for $n=2.4$.  
Moreover, the standard deviations were 0.006 for $n=1.3$ and 0.239 for $n=2.4$, indicating that $n=1.3$ provided faster and more stable transitions, while both $n=1.3$ and $n=2.4$ demonstrated a 100\% success rate in the table-to-wall transition.
One possible explanation is that the steeper curve at $n=2.4$ requires greater pushing force at certain angles, leading to subtle delays during transition.
In contrast, the smoother slope at $n=1.3$ allows for a more uniform distribution of force, enabling quicker and more stable transitions.
Thus, the slope component was designed using $n=1.3$ in equation~(\ref{eq:slope}).  

\subsection{Fabrication}  

\begin{figure}[tbhp]  
    \centering  
    \includegraphics[width=1.0\columnwidth]{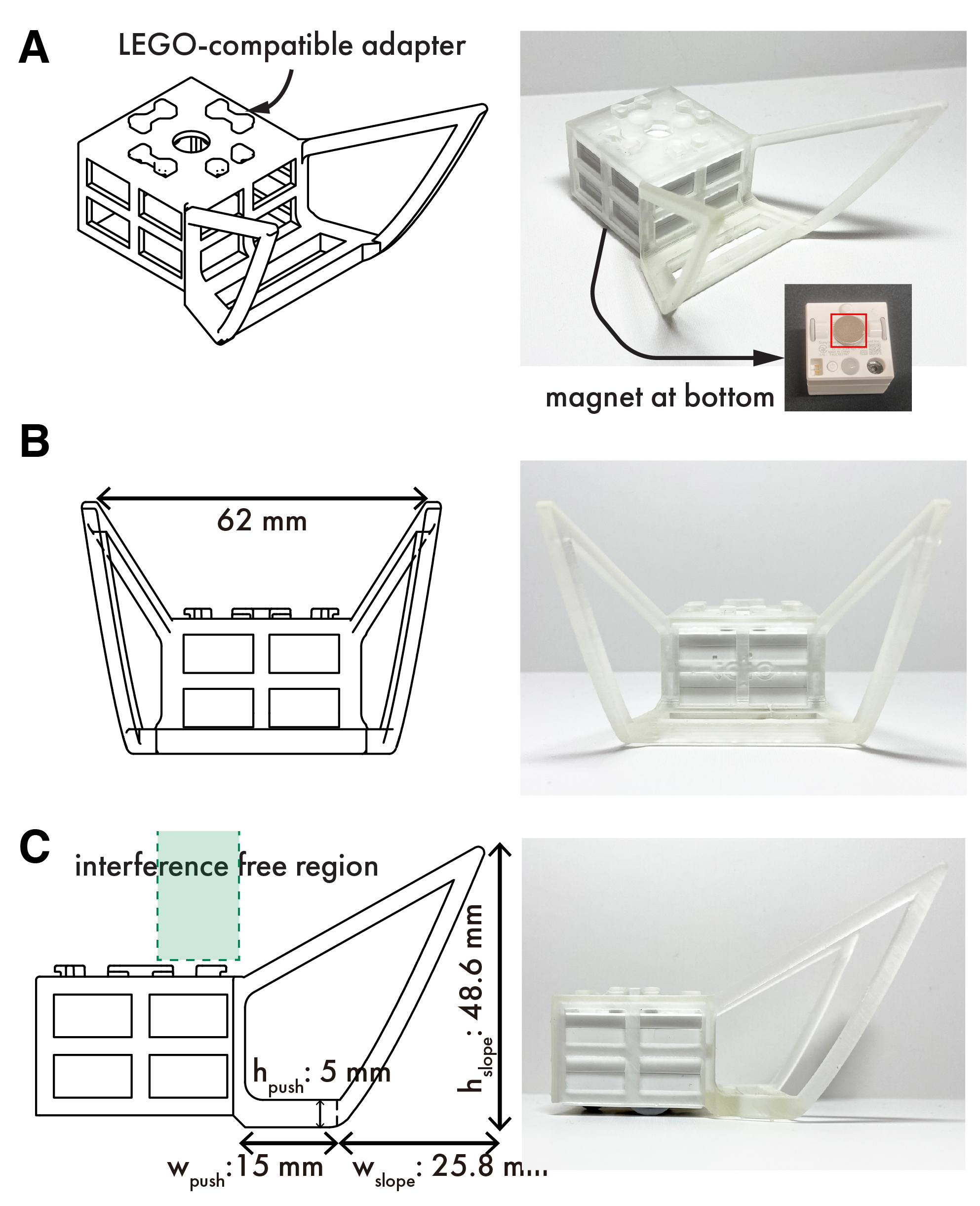}  
    \caption{CAD and photos of the fabricated attachment. (A) Top-right view, (B) front view, and (C) side view.}  
    \Description{CAD and photos of the fabricated attachment. (A) top-right view, (B) front view, and (C) side view. Each view shows the parameters for the final design.}  
    \label{fig:attachment_cad}  
\end{figure}  

Derived from the experimental results, we created a prototype of corobos with the toio robot, as shown in Figure~\ref{fig:attachment_cad}.  
The toio robots have LEGO-compatible adapters on top, so we designed the attachment to directly fit by pressing it onto the robot.  
We also attached a disc-shaped neodymium magnet to the bottom of the robot, similar to previous work~\cite{UbiSwarm2017,HERMITS2020}, to enable the robots to adhere to the wall surface, adding normal force to improve torque while on the tabletop (Figure~\ref{fig:attachment_cad}(A)).  
The magnet had a diameter of 12 $mm$, a thickness of 1.3 $mm$, and a surface magnetic flux density of 0.20 $T$.
LEGO-compatible adapters were added on top for attaching additional objects or functions.  
As illustrated in Figure~\ref{fig:attachment_cad}(C), additional payloads can be placed in the green-filled region, as the assisting robot does not interfere with this area during the transition.  
The attachments were fabricated using an SLA 3D printer (Formlabs Form 3+ with Clear V4 resin).  

To encourage further research and development, we open source the design of the attachment, including the CAD files and fabrication instructions\footnote{\url{https://github.com/hanchangyo/corobos}}.

\subsection{System Setup}  
\begin{figure}[thbp]
    \centering  
    \includegraphics[width=0.8\columnwidth]{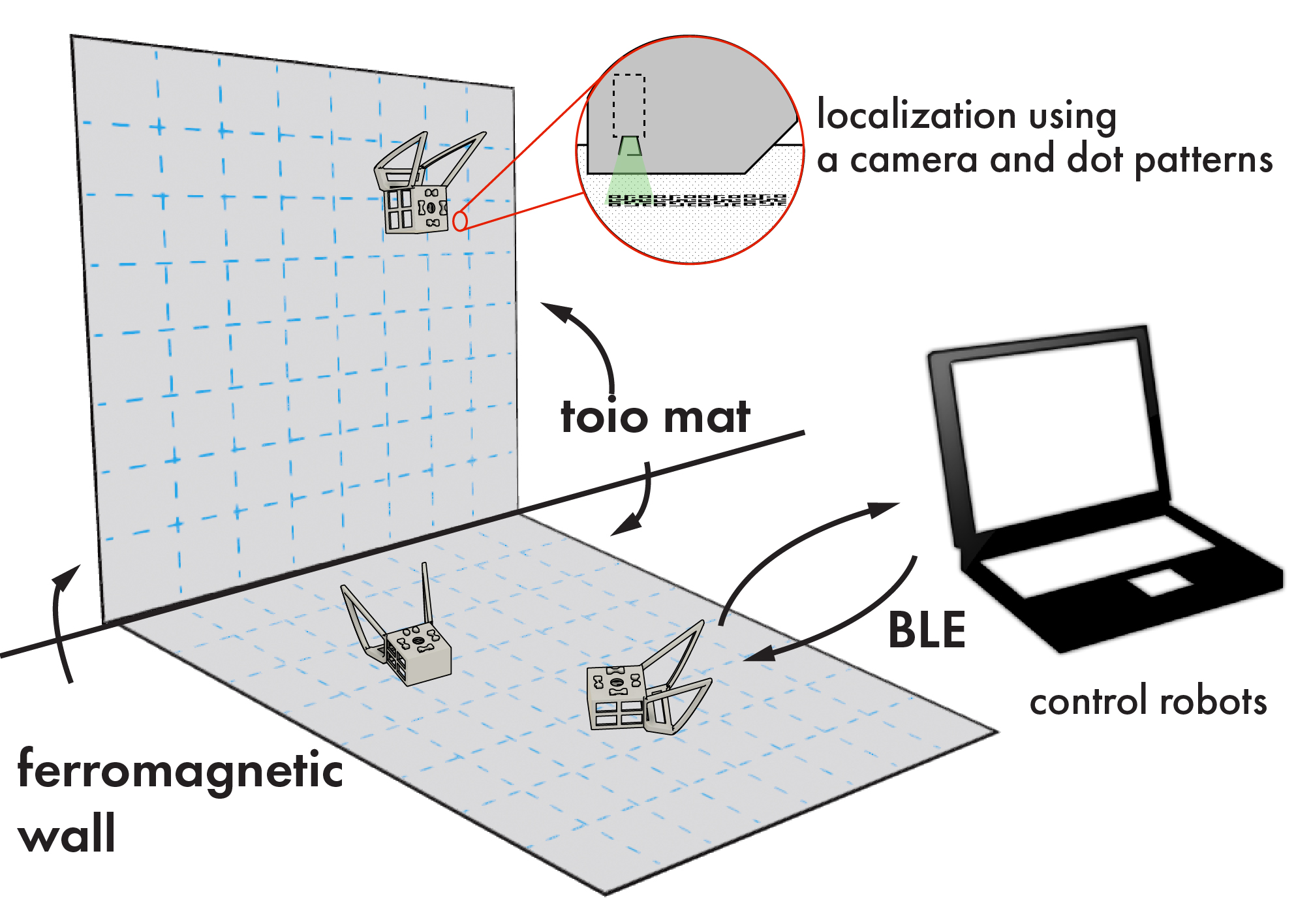}  
    \caption{System setup. toio Mats are placed on the table and the wall. A laptop controls the robots via BLE.}  
    \Description{An illustration of the system setup. On the table and the wall, toio Mats are placed. The robots move on the wall and the table. A laptop communicates with the robots via BLE.}  
    \label{fig:system_config}  
\end{figure}  

Figure~\ref{fig:system_config} illustrates the system setup.  
We developed control software using Unity to manage the robots within the environment.  
We built on the toio SDK for Unity as a foundation and implemented the transition motions as a wrapper function\footnote{\url{https://github.com/morikatron/toio-sdk-for-unity}}.  
The robots are controlled by a laptop (Macbook Pro 2021) via a BLE connection.  
As mentioned in the Experiment section, the position coordinates of the toio robots are obtained from the toio Mat (a specialized paper with micro-dot patterns) using the internal infrared camera of the toio.
We used a regular steel whiteboard as the ferromagnetic surface which had a thickness of 3 $mm$.
The magnetic adhesion force to the ferromagnetic surfaces was approximately 250 $gf$.
The toio Mat was affixed to both the tabletop and wall surfaces to enable accurate tracking of the robots.  

\subsection{Transition Stability Test}  
\begin{figure}[tbhp]  
    \centering  
    \includegraphics[width=1.0\columnwidth]{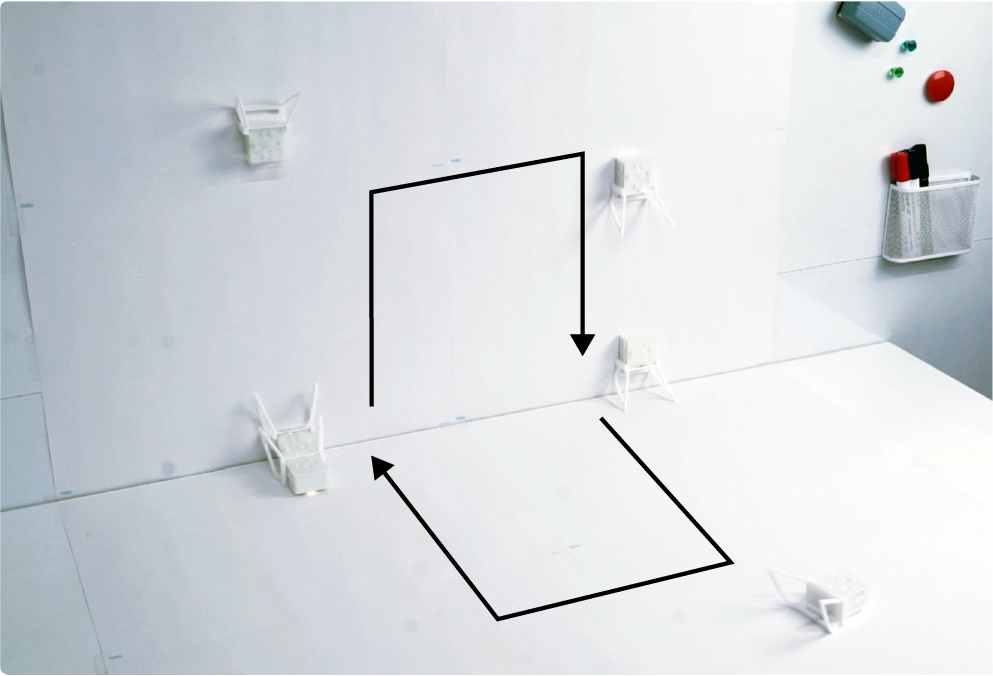}  
    \caption{Transition along the circular route across the table and the wall.}  
    \Description{A route is drawn across the wall and the table, and six robots continuously move along this route.}  
    \label{fig:circular motion}  
\end{figure}  

To evaluate the stability of transitions, we developed a script that directs the robots to follow a predetermined circular route along the tabletop and wall surface.  
With a minimum of three robots, consistent navigation between the tabletop and wall is feasible.  
We employed six toio robots with the attachment to traverse the specified circular route, as shown in Figure~\ref{fig:circular motion}.  
The robots followed this path continuously for nearly two hours until their batteries were depleted, without any transition failures.  
Since each robot has identical functionality, if one fails (e.g., due to a drained battery), another robot can seamlessly take over the role of the failed robot.

\subsection{Payload Test}  

To evaluate the robots' capacity to transport objects between the tabletop and the wall, we conducted a series of payload experiments.  

\subsubsection{Payload Tolerance}  

\begin{figure}[tbhp]  
\centering  
\includegraphics[width=1.0\columnwidth]{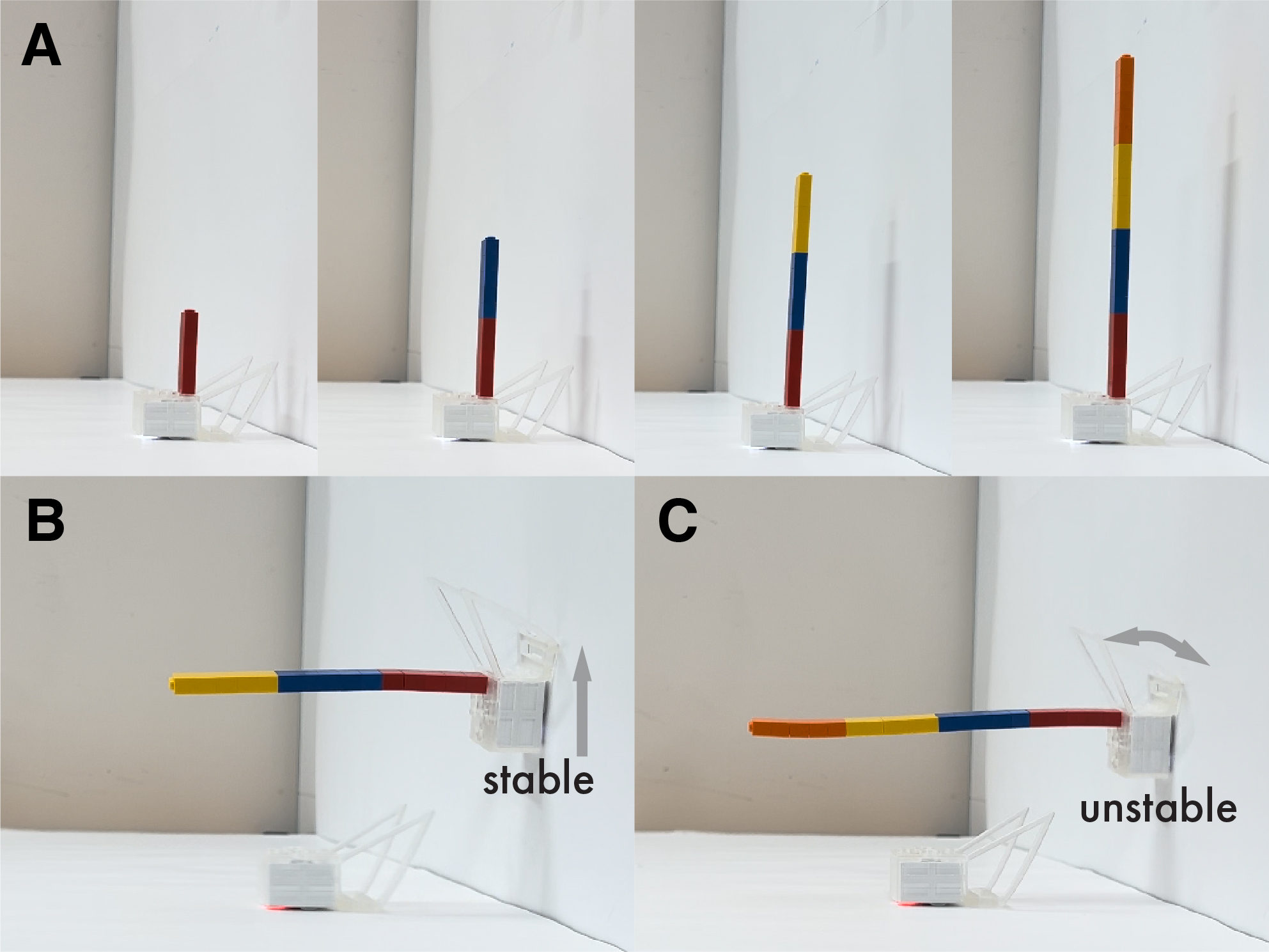}  
\caption{(A) Robots with four incrementally longer rods attached.  
(B) The third rod enables stable transitions and movement on both surfaces.  
(C) The fourth rod leads to transition failures and unstable wall movement due to increased torque.}  
\Description{Robots with various lengths of payload objects attached.}  
\label{fig:payload_length}  
\end{figure}  

We assessed the robots' maximum payload tolerance regarding length and weight.  
On the wall, gravitational forces affect the robot's maneuverability when carrying a payload, limiting its capacity based on the length and torque of the attached object.  
Similarly, on the tabletop, while weight has less impact, increased payload height can destabilize the robot by raising its center of mass.  

Figure~\ref{fig:payload_length}(A) shows four payload rods constructed from 1 x 2 LEGO bricks, each extended in increments of five bricks.  
Each increment weighed 4 grams and measured 43 mm in height.  
After 50 transition trials, stability was maintained up to the third rod increment (Figure~\ref{fig:payload_length}(B)).  
However, the fourth increment frequently failed during transitions and exhibited unstable movement on the wall due to the increased torque (Figure~\ref{fig:payload_length}(C)).  
Therefore, the robots can tolerate a payload torque of approximately 0.0076 $N \cdot m$ on both surfaces up to the third increment.  

\subsubsection{Example Objects}  

\begin{figure}[tbhp]  
\centering  
\includegraphics[width=1.0\columnwidth]{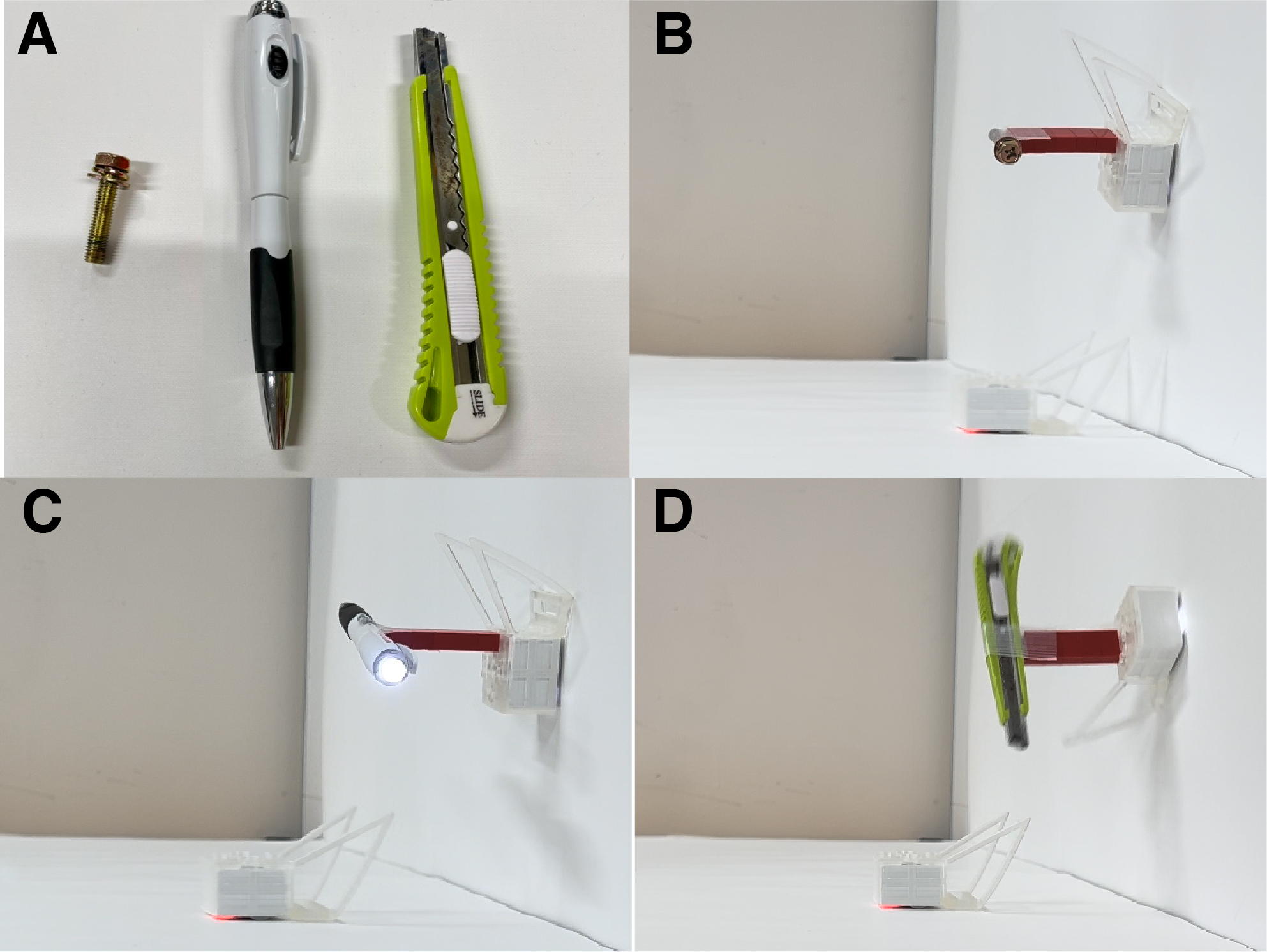}  
\caption{(A) Three objects tested for transition and movement stability:  
(B) a bolt, (C) an LED penlight, and (D) a cutter knife.}  
\Description{Three different objects used in the payload test: a bolt, an LED penlight, and a cutter knife.}  
\label{fig:payload_weight}  
\end{figure}  

To demonstrate the range of objects the robots can transport, we tested three common items used in table work (Figure~\ref{fig:payload_weight}(A)):  
a bolt, an LED penlight, and a cutter knife, weighing 9, 15, and 20 grams, respectively.  
These objects were secured to the end of a five-brick LEGO assembly using adhesive tape.  
Each item underwent 50 transition tests between surfaces, successfully maintaining stable transitions and movements (Figure~\ref{fig:payload_weight}(B)(C)(D)).  
With this capability, the robots can deliver items from one user's table to another by traversing the wall.  

\section{Application Examples}\label{sec:applications}
We showcase the versatility of \textit{corobos} through various application scenarios.
Drawing on our design implications, each example leverages the spatial interplay between accessible (table) and peripheral or out-of-reach (wall) surfaces to shape user attention, engagement, and collaboration along with surface transition.
This movement encourages users to shift between focused manipulation on the table and broader awareness on the wall, thereby opening new interaction design possibilities.
We annotate the elements of the design space used in the examples to illustrate their usage, as summarized in Table~\ref{tab:design_space}.

While prior works such as \textit{ThrowIO}~\cite{ThrowIO_CHI2023} and \textit{(Dis)Appearables}~\cite{disappearables2022} explore multi-surface interactions, they differ in their approach to defining and utilizing interactive spaces.
\textit{ThrowIO} focuses solely on ceiling-based interactions, leveraging the unique properties of overhanging surfaces to enable throwing and catching interactions.
The ceiling serves as the primary interaction plane, allowing robots to manipulate objects in mid-air and drop them to users below.
In contrast, \textit{(Dis)Appearables} introduces two distinct spaces--front-stage and back-stage--where robots dynamically transition between visible and hidden states to control the appearance and disappearance of mobile robots.
This approach emphasizes visibility-based interaction.
Our work, \textit{corobos}, also utilizes two distinct surfaces but differentiates them based on user reachability rather than visibility.
Both surfaces are visible to users but the utilization of the surfaces are differed by the reachability.

\begin{table*}[h]
\centering
\caption{Design space components used in each application example}
\label{tab:design_space}
\begin{resizebox}{1.0\textwidth}{!}{

\begin{tabular}{|l|ccc|cccc|cccc|cccc|}
\hline
\multirow{3}{*}{Application Examples} & \multicolumn{7}{c|}{\textbf{Payload Objects}} & \multicolumn{4}{c|}{\textbf{User Manipulation}} & \multicolumn{4}{c|}{\textbf{Robot Capabilities}} \\
\cline{2-16}
& \multicolumn{3}{c|}{Attachment Method} & \multicolumn{4}{c|}{Object Types} & & & & & & & & \\
\cline{2-16}
& \rotatebox{90}{Rigid Fixtures} & \rotatebox{90}{Tethered} & \rotatebox{90}{Pushed} & \rotatebox{90}{Small Items} & \rotatebox{90}{Surface-Adhered Items} & \rotatebox{90}{Papers} & \rotatebox{90}{Tangibles} & \rotatebox{90}{Attach/Detach} & \rotatebox{90}{Touch} & \rotatebox{90}{Move} & \rotatebox{90}{GUI} & \rotatebox{90}{Transporting Payloads} & \rotatebox{90}{Reallocating Robots} & \rotatebox{90}{Synchronizing Movements} & \rotatebox{90}{Creating Spatial Displays} \\
\hline
Organizing Workspace & \checkmark & & & \checkmark & & & & \checkmark & \checkmark & & & \checkmark & & & \\
\hline
Delivering Heavy Objects & & & \checkmark & & \checkmark & & & & & & & & \checkmark & & \\
\hline
Dynamic Wall Posting & \checkmark & & & & & \checkmark & & \checkmark & \checkmark & \checkmark & & \checkmark & & \checkmark & \\
\hline
Educational Tools & \checkmark & & & & & & \checkmark & \checkmark & \checkmark & & & \checkmark & \checkmark & & \\
\hline
Emulating Room Layout & \checkmark & & & & & & \checkmark & \checkmark & & & \checkmark & \checkmark & & & \checkmark \\
\hline
Spatial Display & & \checkmark & & & & & \checkmark & \checkmark & \checkmark & & &  & & & \checkmark \\
\hline
Workshop Brainstorming Support & \checkmark & & & & & \checkmark & & \checkmark & \checkmark & \checkmark & & \checkmark & & \checkmark & \\
\hline
\end{tabular}
}
\end{resizebox}
\end{table*}

\subsection{Organizing Workspace}\label{app:workspace}
\begin{figure}[tbhp]
    \centering
    \includegraphics[width=1.0\columnwidth]{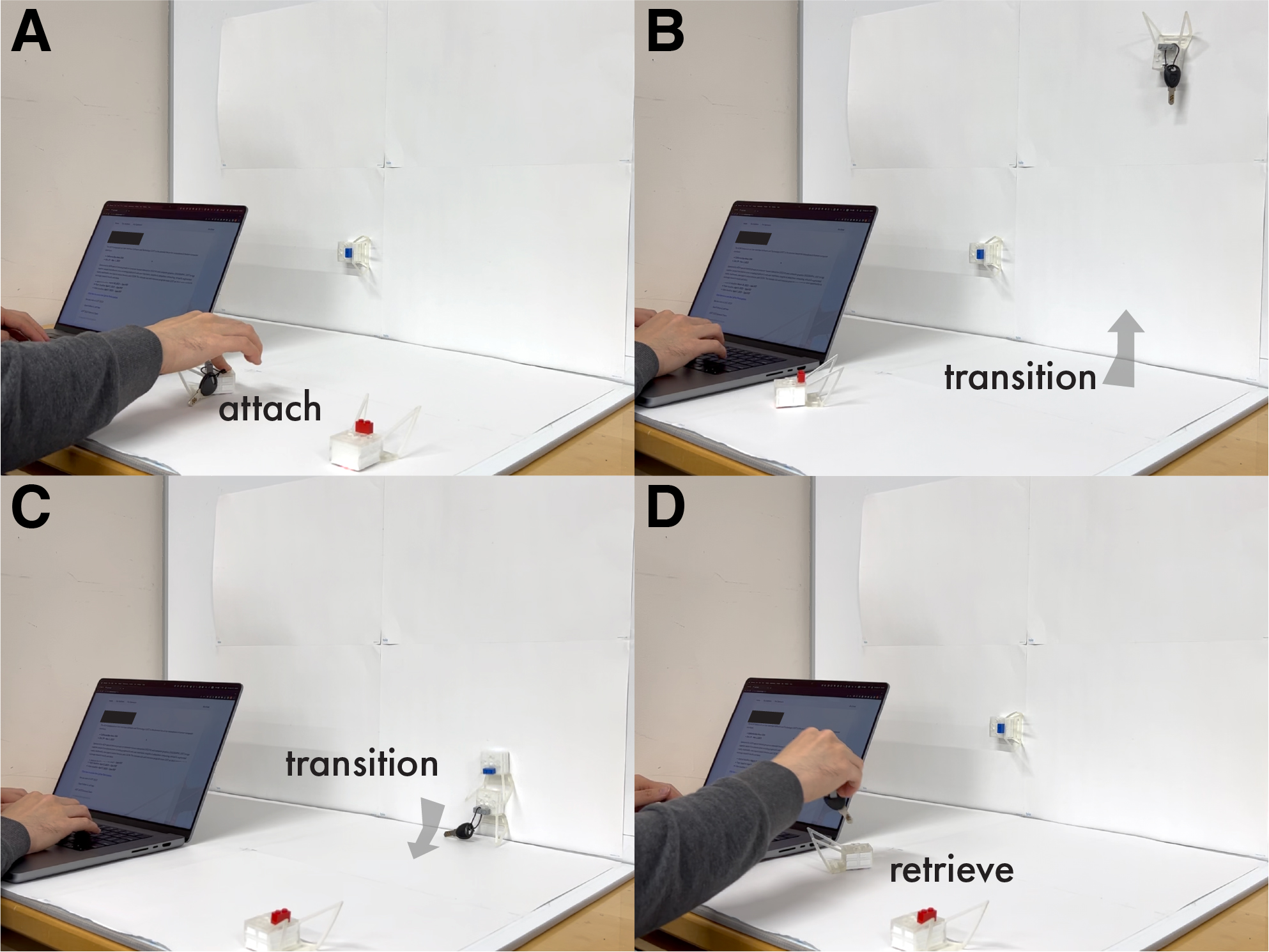}
    \caption{Robots help users organize the workspace by moving redundant items to the wall. (A) The user attaches a key to the robot, (B) the robot transitions from the table to the wall with the key and stores it on the wall. (C) When the user requests the key, the robot descends and (D) delivers the key to the user.}
    \Description{A robot brings a key to the wall for storage. When the user requests the key, the robot brings it back to the table.}
    \label{fig:workspace_organization}
\end{figure}

\textit{corobos} effectively helps organize workspaces by moving redundant items to the wall.
In this scenario, the user attaches a key (approximately 8 grams) to a robot on the table~(Figure~\ref{fig:workspace_organization}(A)).
The robot transitions from the tabletop to the wall, placing the key in a designated location out of the user's immediate reach (Figure~\ref{fig:workspace_organization}(B)).
When the user needs the key, a tap on the robot on the table prompts it to retrieve the key from the wall and deliver it back to the user~(Figure~\ref{fig:workspace_organization}(C)(D)).
This interplay leverages spatial transitions to emphasize which objects warrant users’ direct, manual engagement, as opposed to those suited for peripheral awareness.

\subsection{Delivering Heavy Objects}\label{app:heavy_object}
\begin{figure}[tbhp]
    \centering
    \includegraphics[width=1.0\columnwidth]{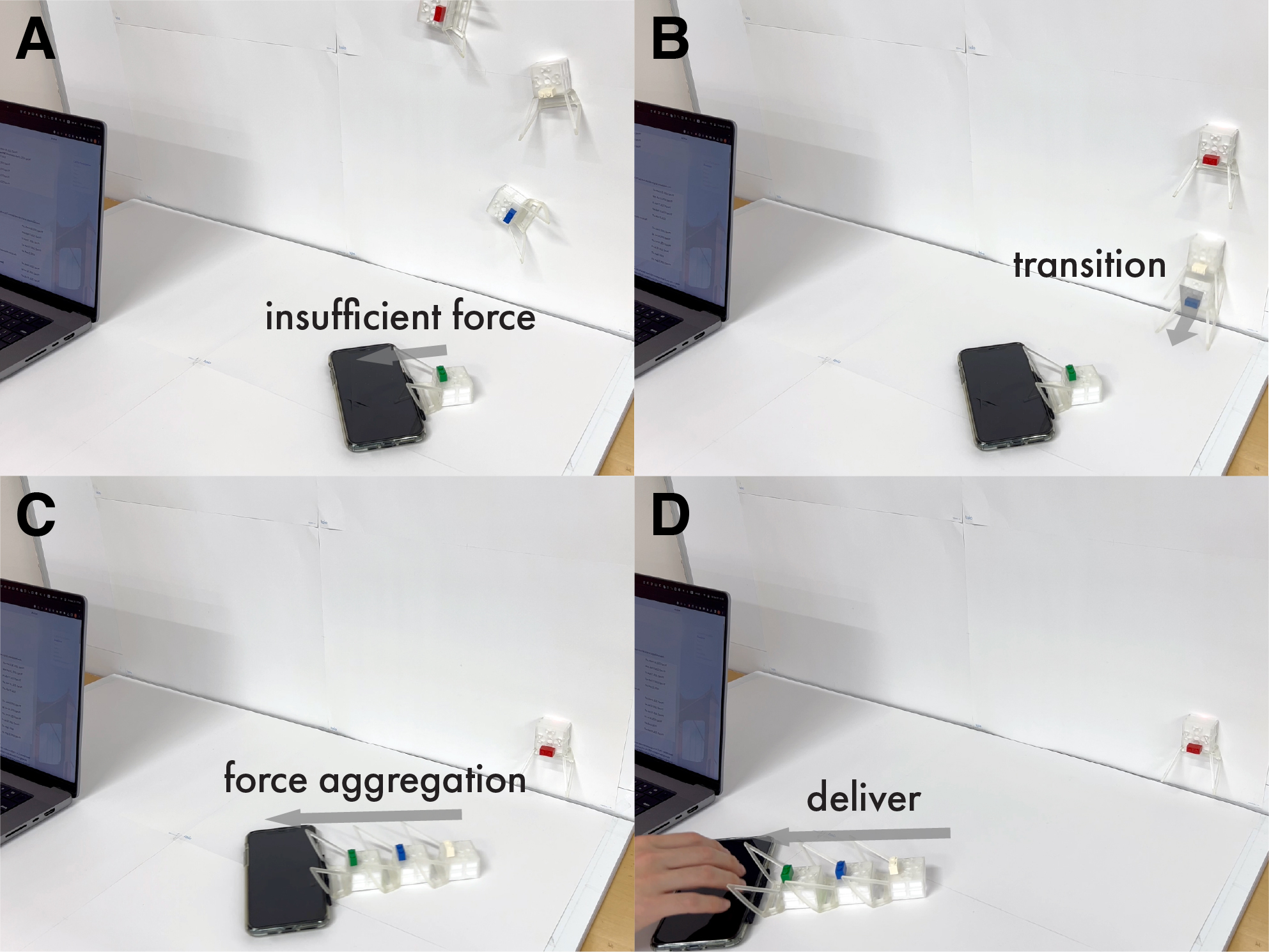}
    \caption{Robots transition to aggregate force to move heavy objects on tabletop. (A) A single robot is unable to push the phone to the user. (B) Two additional robots descend from the wall and (C) push the phone together (D) to deliver it to the user for an important notification.}
    \Description{A robot tries to push a smartphone on the table but fails. Two robots descend from the wall to help, and together they push the phone to the user.}
    \label{fig:move_heavy_objects}
\end{figure}

\begin{figure}[tbhp]
    \centering
    \includegraphics[width=1.0\columnwidth]{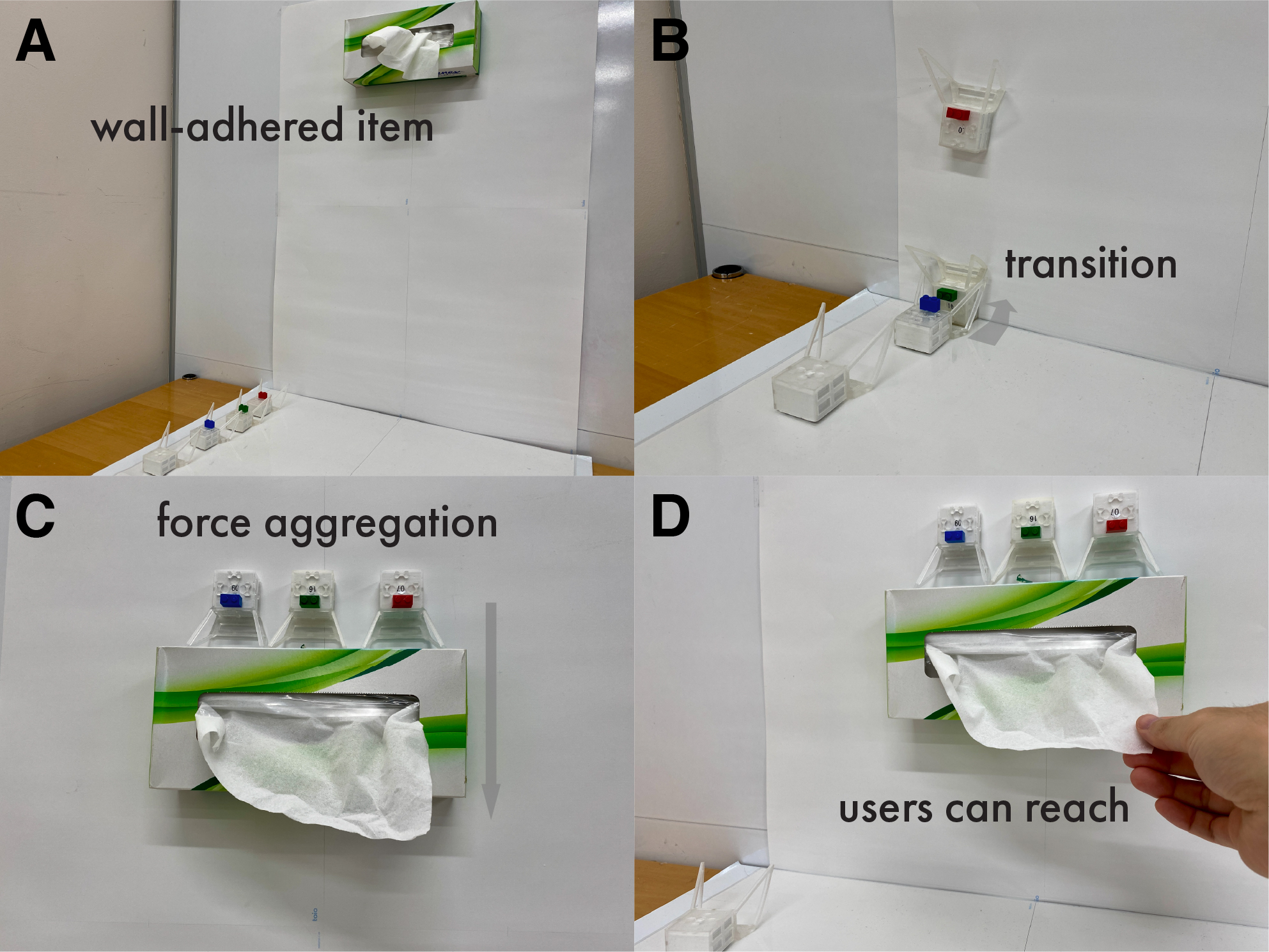}
    \caption{Robots transition to aggregate force to move heavy objects on the wall. (A) A wall-mounted tissue paper box is located too high for users to reach. (B) Three robots ascend the wall and (C) push the tissue paper box downward (D) to move it to a lower, accessible position.}
    \Description{A tissue paper box is located high on the wall. Three robots ascend the wall and push the box downward. A user's hand picks up tissue paper from the box.}
    \label{fig:move_wall_heavy_objects}
\end{figure}

Robots can remain on the wall to optimize desk space, keeping the workspace clear.
When a specific number of robots is needed for a task on a particular surface, they can transition between surfaces as required.

In one scenario, a robot on the table tries to deliver a smartphone to the user to notify them of an important message, but the device is too heavy (approximately 228 grams) for a single robot (Figure~\ref{fig:move_heavy_objects}(A)).
Robots stationed on the wall can descend to assist, providing combined force to deliver the phone for the notification (Figure~\ref{fig:move_heavy_objects}(B)(C)(D)).

This scenario exemplifies how spatial separation guides user attention:
Rather than clutter the table with unused robots, they remain on the wall until summoned, thus aligning the immediate workspace with the user’s current priorities.

In another example, a tissue paper box is stored high on the wall where users cannot easily reach it, optimizing wall space usage (Figure~\ref{fig:move_wall_heavy_objects}(A)). Several robots transition from the table to the wall~(Figure~\ref{fig:move_wall_heavy_objects}(B)) and push the tissue paper box downward to a more accessible position~(Figure~\ref{fig:move_wall_heavy_objects}(C)). Users can then easily retrieve tissue paper from the box~(Figure~\ref{fig:move_wall_heavy_objects}(D)).

\subsection{Dynamic Wall Posting}\label{app:wall_object}
\begin{figure}[tbhp]
    \centering
    \includegraphics[width=1.0\columnwidth]{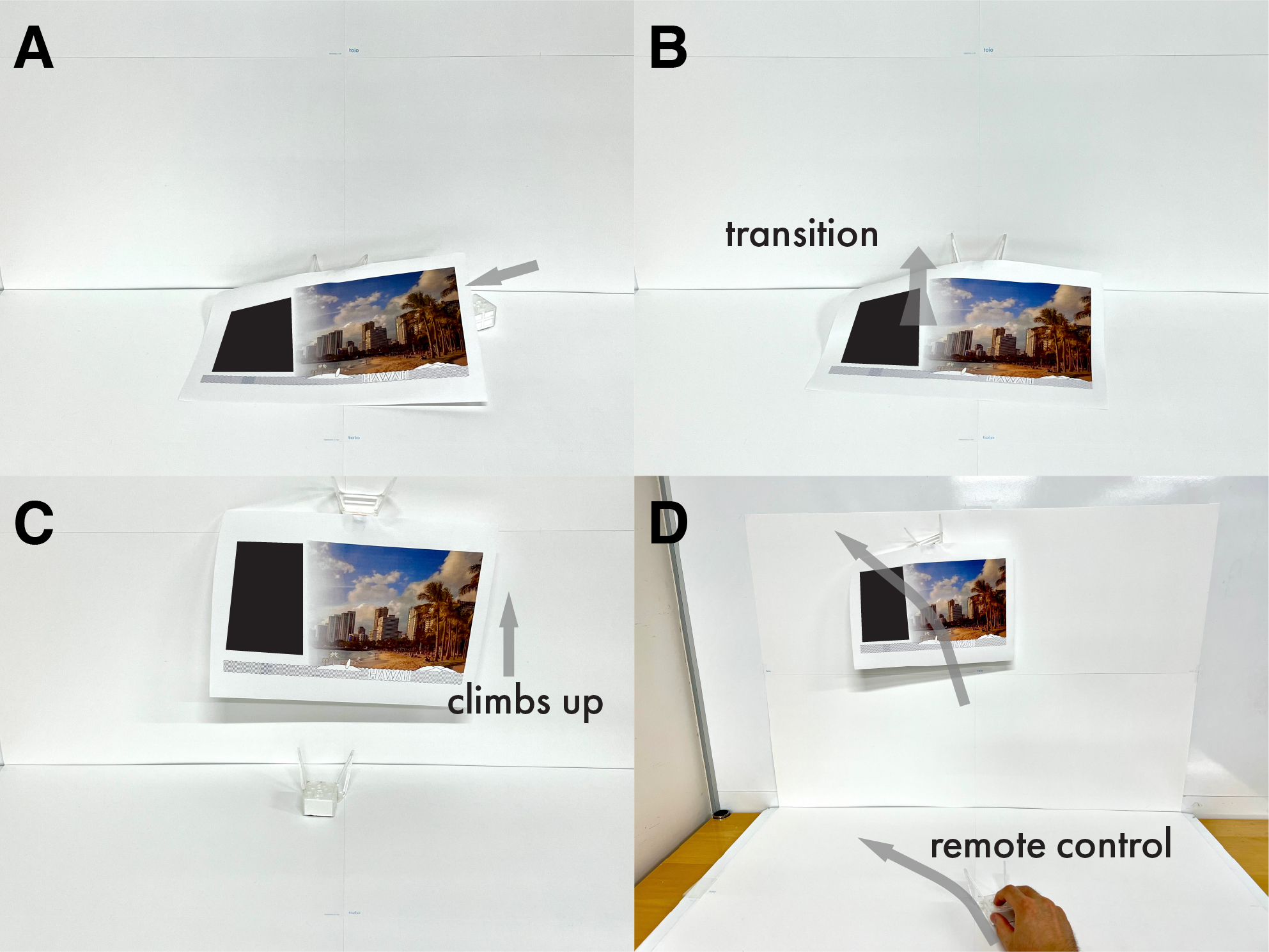}
    \caption{Manipulating wall objects using the transportation and synchronization feature. (A) The user attaches a CHI poster to a robot on the table, then a helper robot moves below the poster. (B) The poster robot transitions to the wall. (C) The poster robot climbs up the wall to a position out of the user's reach and (D) is remotely controlled by a robot on the table to adjust positioning.}
    \Description{The user attaches a poster labeled 'CHI' to a robot on the table. The robot ascends the wall, and the user adjusts its position by manipulating another robot on the table.}
    \label{fig:dynamic_wall}
\end{figure}

\textit{corobos} can move and manipulate wall objects remotely.
This application demonstrates this capability by positioning a poster on the wall.
The user attaches the poster (approximately 1 gram) to a robot~(Figure~\ref{fig:dynamic_wall}(A)) on the table for setup, which works with another robot to transport it to the wall~(Figure~\ref{fig:dynamic_wall}(B)(C)).
The robot on the table then acts as a remote control for the one on the wall, synchronizing their movements.
The user can use the table robot to remotely move the poster to the desired position~(Figure~\ref{fig:dynamic_wall}(D)).
This shifts users’ focus from intimate, manual interactions (table) to broader, visually oriented observation (wall).

However, \textit{corobos} must be carefully controlled to ensure that thin and large objects, such as posters, do not obstruct robot movements.
If a large piece of paper is positioned in the path of an assisting robot, it may prevent successful interaction.
This limitation highlights the need for planned robot navigation and spatial awareness to avoid obstacles when transporting such objects.

\subsection{Educational Tools}\label{app:education}

\begin{figure}[tbhp]
    \centering
    \includegraphics[width=1.0\columnwidth]{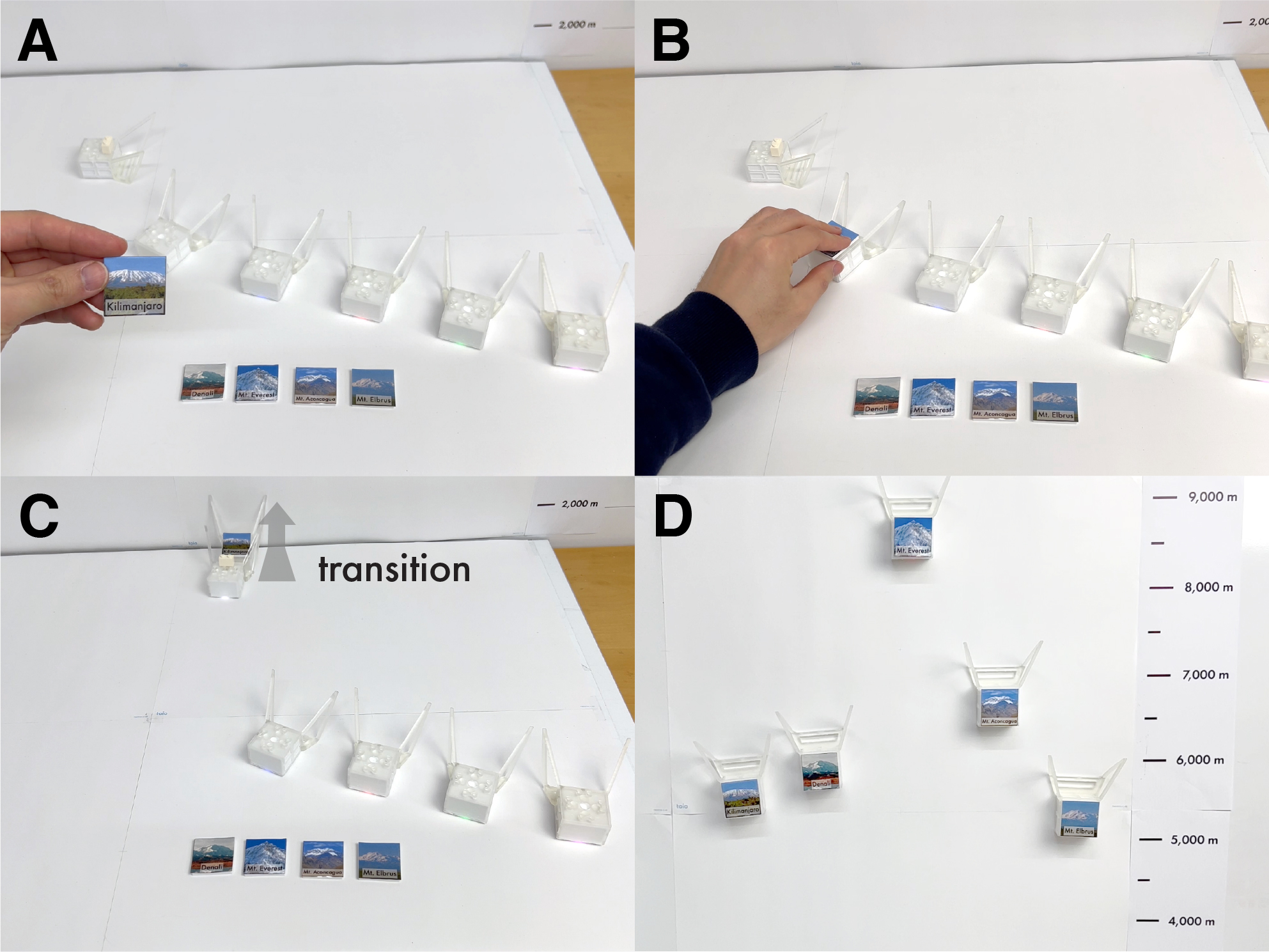}
    \caption{An educational tool example to visually demonstrate the elevations of the world's highest peaks.
    (A) The user attaches a panel featuring an illustration of one of the world's highest peaks to the robot and (B) taps the robot for activation.
    (C) The robot transitions to ascend the wall and (D) climbs to the corresponding height of the peak.}
    \Description{The user attaches a panel featuring an illustration of one of the world's highest peaks to the robot.
    The robot then ascends the wall to represent the corresponding height of the peak.
    The height label is displayed on the right side of the figure.}
    \label{fig:educational tools}
\end{figure}

\textit{corobos} can serve as an educational tool by using its ability to move from the table to the wall, representing vertical height.
In this scenario, the robot’s initial position on the table symbolizes a familiar baseline, helping learners intuitively grasp the concept of ascending to higher elevations.
We demonstrate how to teach children about the world's highest peaks using the robot's internal hall sensor, which detects nearby magnetic fields.
When the user attaches a magnetic panel labeled with a peak’s name to the robot and taps it on the table~(Figure~\ref{fig:educational tools}(A)(B)), the robot ascends the wall.
It then stops at the height corresponding to that peak~(Figure~\ref{fig:educational tools}(C)(D)).
This transition from reachable (table) to distant (wall) surfaces embodies the abstract notion of height, encouraging learners to connect physical movement with cognitive understanding.

\subsection{Emulating Room Layout}\label{app:room_layout}

\begin{figure}[tbhp]
    \centering
    \includegraphics[width=1.0\columnwidth]{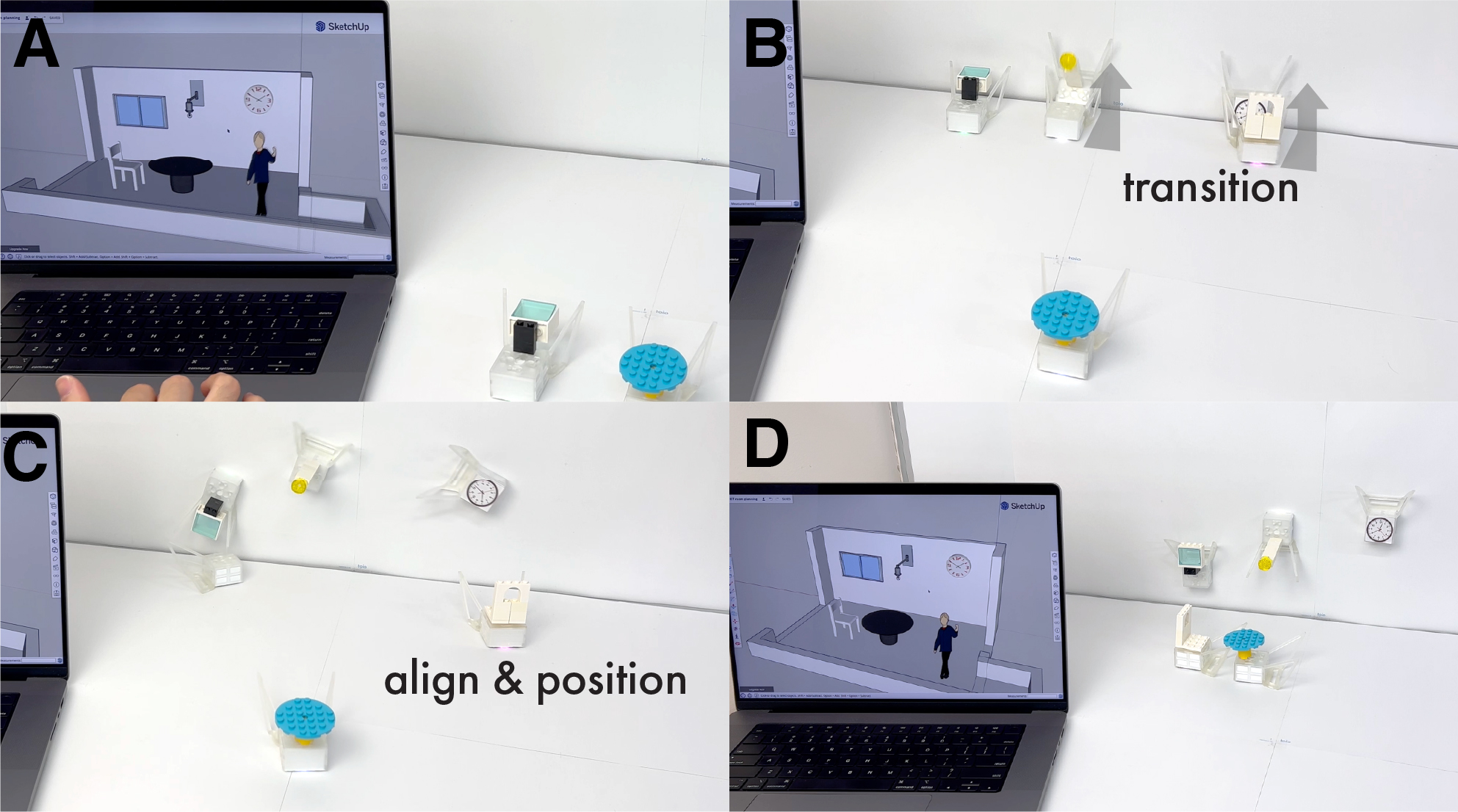}
    \caption{(A) A CAD software displays a room layout with furniture (table, chair, wall light, wall clock, and window).
    (B) The robots, equipped with attachments representing each piece of furniture, transition to the wall and (C)(D) emulate the layout as shown on the screen.}
    \Description{There is a layout of a room on the laptop display.
    Robots are equipped with attachments representing each piece of furniture in the room layout.
    They move on the wall to emulate the layout shown on the screen.}
    \label{fig:room_plan}
\end{figure}

\textit{corobos} can help users emulate room layouts by first allowing them to work hands-on at the table, where all manual adjustments are more accessible.
Users begin by designing a layout in CAD software~(Figure~\ref{fig:room_plan}(A)) and then place furniture proxies on the tabletop, treating it as a familiar ``floor plan'' workspace.
Since the table is easy to reach and manipulate, users can freely experiment with positioning items before making final decisions.
Once satisfied with the arrangement, robots transition selected items to the wall, representing higher or out-of-reach elements in the room~(Figure~\ref{fig:room_plan}(B)(C)(D)).
By moving from a user-accessible horizontal surface to a vertical display surface, \textit{corobos} aids in bridging conceptual planning with real-world spatial arrangements, helping users visualize both the overall room configuration and hard-to-reach features in a tangible, incremental manner.

\subsection{Spatial Display}\label{app:spatial_display}

\begin{figure}[tbhp]
    \centering
    \includegraphics[width=1.0\columnwidth]{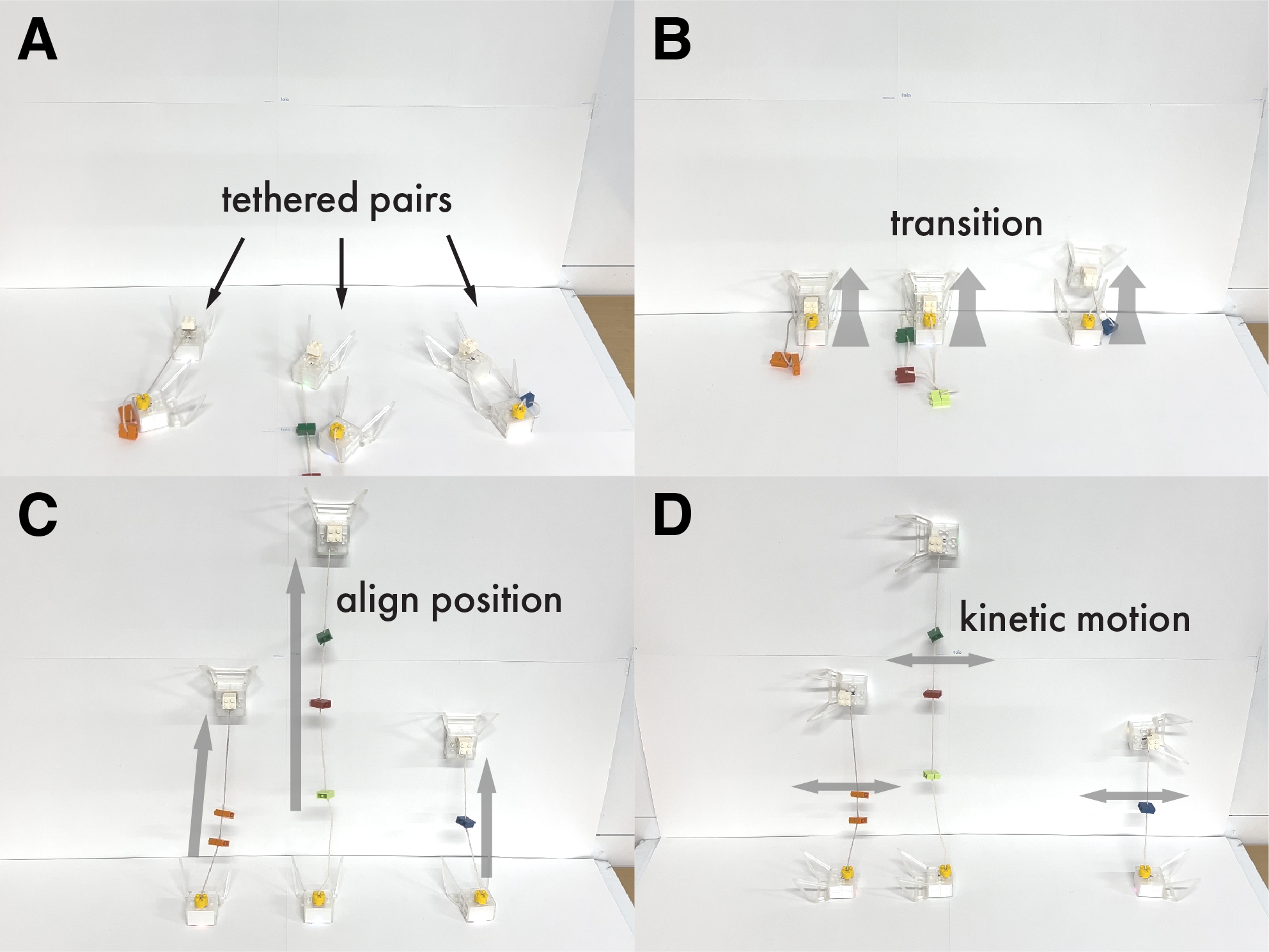}
    \caption{(A) Three robot pairs are tethered with strings, and tangibles are placed in the middle.
    (B) One robot from each pair transitions to the wall, (C) causing the tangibles to float mid-air, and (D) enabling the robots to impart kinetic motion.}
    \Description{Three pairs of robots start on the table, each tethered by a string.
    One robot from each pair moves to the wall, creating a suspended line between the wall and table.}
    \label{fig:string_display}
\end{figure}

\textit{corobos} can create spatial displays that leverage both the accessible tabletop and the vertically oriented wall to form dynamic, three-dimensional scenes.
Robots begin on the table, where users can easily attach tangibles or adjust configurations~(Figure~\ref{fig:string_display}(A)).
When one robot from each pair transitions to the wall~(Figure~\ref{fig:string_display}(B)), it pulls the tethered string taut, suspending tangibles in mid-air~(Figure~\ref{fig:string_display}(C)).
This transforms flat layouts into vertical, layered representations that users can observe from various angles.
Because the wall is more distant and serves as a display surface, these suspended tangibles can represent hierarchical data, trends over time, or other structured information.
The robots can even move to create kinetic effects~(Figure~\ref{fig:string_display}(D)), making the displayed information more engaging.
By bridging the gap between a hands-on, easily modifiable table and a visually prominent wall display, \textit{corobos} supports more meaningful and memorable interactions with complex spatial data.

\subsection{Workshop Brainstorming Support}\label{app:workshop}

\begin{figure}[tbhp]
    \centering
    \includegraphics[width=1.0\columnwidth]{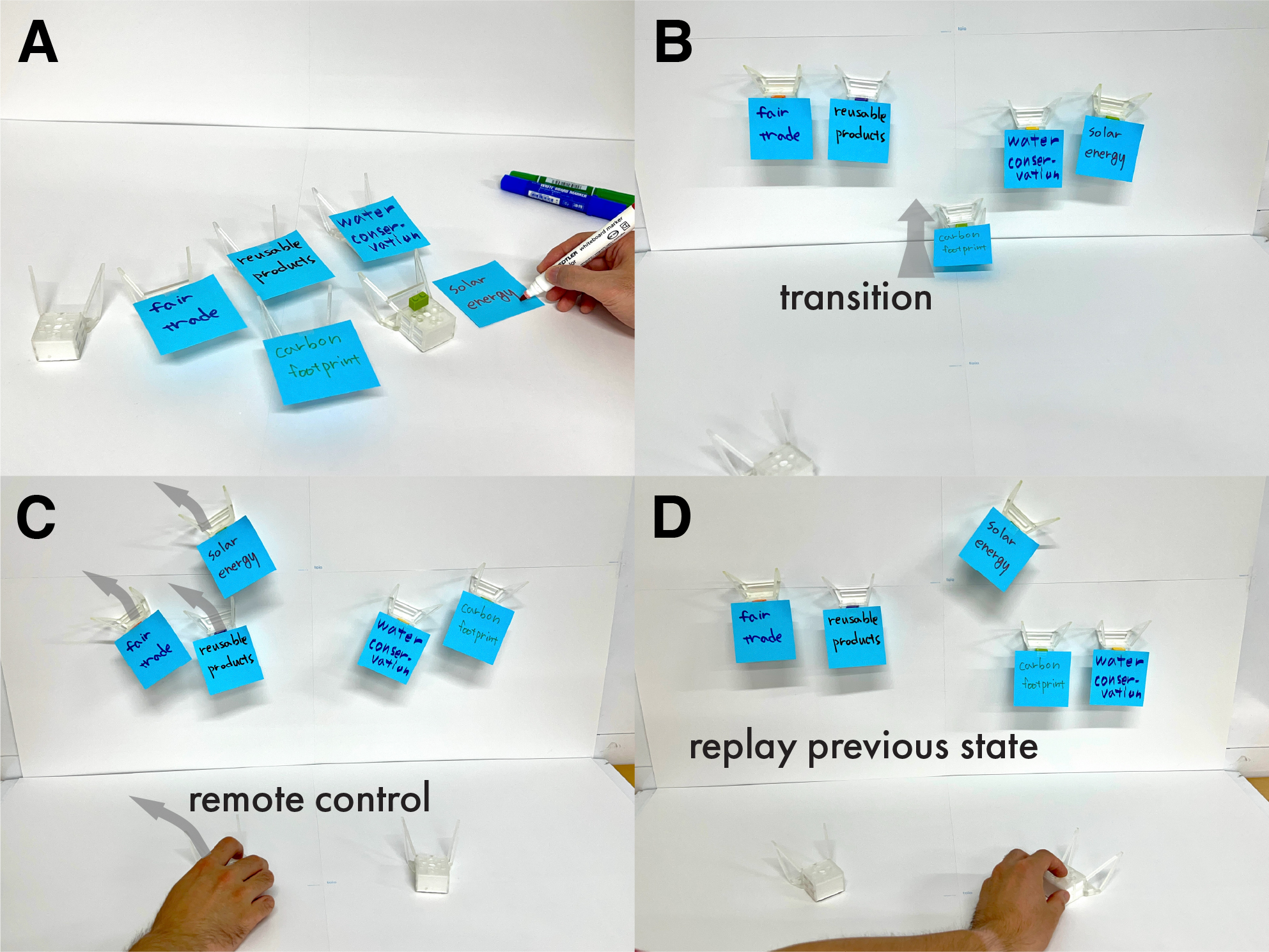}
    \caption{Robots are used to display ideas during a workshop session.
    (A) A participant writes ideas on Post-it notes and attaches them to robots on the table.
    (B) The robots ascend the wall to present the ideas.
    (C) During discussion, participants move and regroup notes using the robots on the table.
    (D) Arrangements can be recorded and revisited to summarize the workshop.}
    \Description{Robots carrying Post-it notes with ideas move from the table to the wall.
    Participants rearrange these notes using the robots on the table.
    The robots can record arrangements for later review.}
    \label{fig:workshop_feedback}
\end{figure}

In a workshop setting, \textit{corobos} enhances brainstorming by letting participants start their idea generation at the table, where writing and attaching Post-it notes to robots is straightforward~(Figure~\ref{fig:workshop_feedback}(A)).
When instructed, the robots ascend the wall, placing ideas in a visually accessible, shared space beyond easy reach~(Figure~\ref{fig:workshop_feedback}(B)).
This vertical reallocation keeps the table clear for ongoing hands-on work, while the wall provides a more collective display of contributions.
Participants can reorganize these ideas by manipulating robots on the table, clustering related concepts or themes~(Figure~\ref{fig:workshop_feedback}(C)).
Moreover, \textit{corobos} can record and later replay previous note arrangements, enabling the group to revisit earlier stages of the discussion for summaries or retrospective analysis~(Figure~\ref{fig:workshop_feedback}(D)).
By integrating easily accessible, hands-on manipulation at the table with the wall’s capacity for broader visibility and retrospective review, \textit{corobos} supports dynamic, collaborative brainstorming that persists beyond the immediate session.

\section{Discussion and Future Work}

\subsection{Transition between Wall--Ceiling and Wall--Wall}
In the current implementation of \textit{corobos}, the robots can transition between tabletop and wall surfaces. 
As a future extension, we aim to explore transitioning between wall and ceiling surfaces, further increasing the system's versatility and adaptability. 
This capability could enable ceiling-based interactions, as presented in \cite{AeroRigUI_CHI2023,ThrowIO_CHI2023,ThreadingSpace2024}. 
However, this development would require overcoming challenges related to gravity and adherence.
A wall-to-wall transition also faces challenges with the transitioning robot potentially falling, as there is no support once the robot detaches from the original surface.

We have considered a prototype using two permanent magnets at the rear of the robot and an electromagnet at the front for wall and ceiling transitions. 
However, several issues remain unresolved, such as the risk of falling and the dual attraction of permanent magnets to ferromagnetic surfaces. 
Future research will focus on designing improved solutions to address these challenges.

\subsection{Rationale Against Using Ramps}
\begin{figure}[tbhp]
    \centering
    \includegraphics[width=1.0\columnwidth]{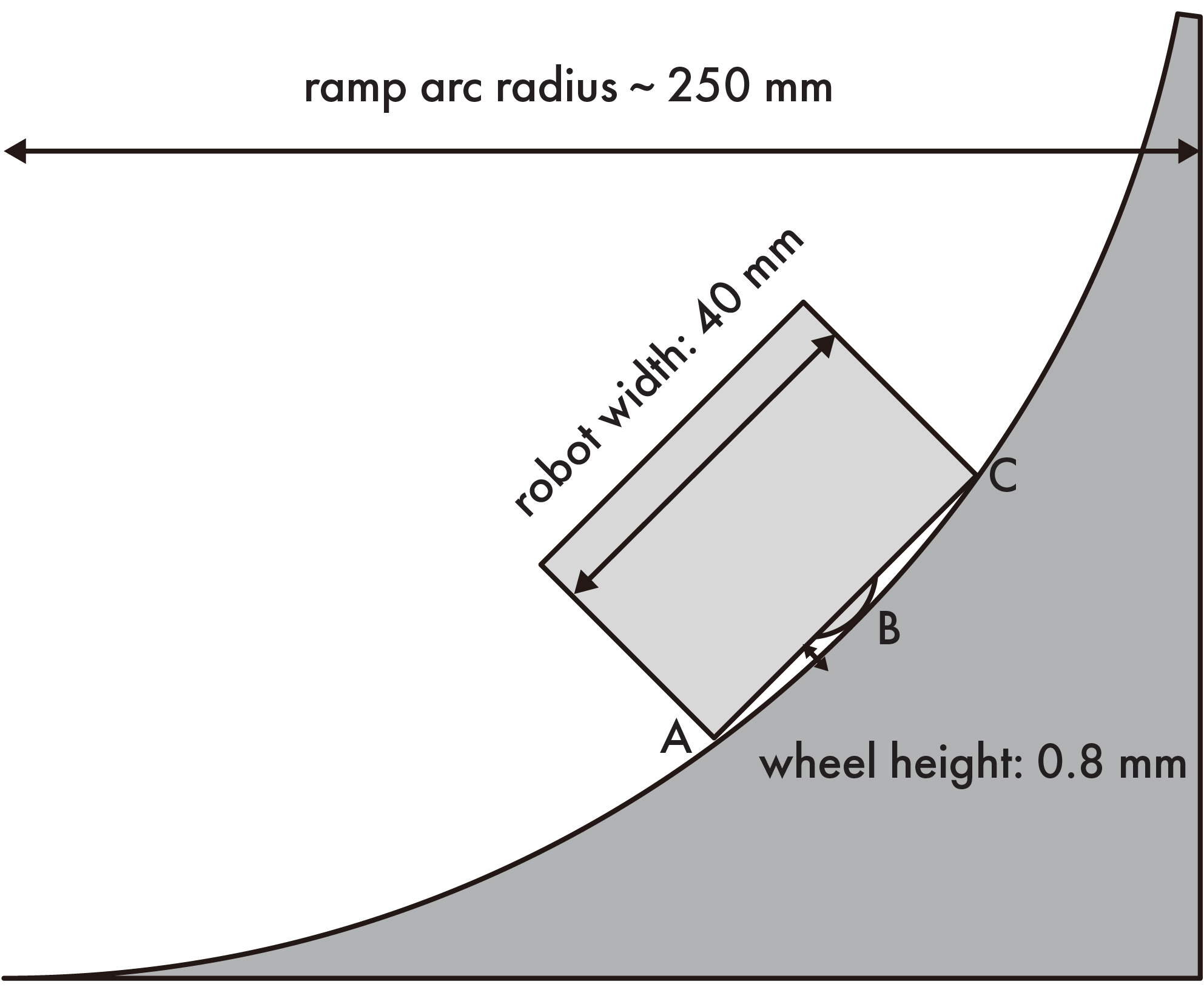}
    \caption{Graphical representation of the required ramp arc radius.}
    \Description{}
    \label{fig:ramp}
\end{figure}

One possible alternative for enabling robots to transition between surfaces is to use a ramp structure on the tabletop.
While this method provides a continuous path for movement, it introduces significant spatial constraints.
Assuming that a cube-shaped robot's body is raised 0.8 mm above the bottom of its wheels and located at the center of the robot, and the width of the robot is 40 mm, a ramp would need to account for this height difference to avoid instability.
A geometric calculation reveals that a ramp with an arc radius of 250 mm would be required to ensure a smooth transition as shown in Equation~\ref{eq:arc}.

\begin{equation}
    R = \frac{w_{robot}^2}{8h_{wheel}}
    \label{eq:arc}
\end{equation}

This equation assumes that the ramp's arc passes through three key points, A, B, and C, as illustrated in Figure~\ref{fig:ramp}.
A primary advantage of using a ramp is that it allows robots to maintain contact with a surface at all times, reducing the need for precise adhesion mechanisms or complex cooperative behaviors.
However, the major drawback is its space inefficiency.
A ramp with the required curvature would occupy a substantial portion of the tabletop, potentially obstructing workspace functionality and limiting user interaction with other objects.
Additionally, once installed, ramps impose fixed transition points, reducing flexibility in robot movement and adaptation to dynamic environments.

In contrast, our design addresses these issues by enabling robots to transition between surfaces autonomously, without requiring permanent physical modifications to the workspace.
This approach maximizes spatial efficiency, allowing robots to operate without consuming excessive tabletop space while maintaining adaptability for various applications.
However, our method relies on precise alignment and cooperation between robots, which may introduce challenges in environments with uneven surfaces or unexpected obstacles.
Future work could explore hybrid approaches, combining small modular ramps with cooperative robot transitions, to balance efficiency and adaptability.

\subsection{Safety Concerns}
Safety is a critical aspect of \textit{corobos}, especially when the robots are in motion and transitioning between surfaces. 
There is a risk of robots falling off the wall surface and potentially causing harm to users. 
Future work will focus on mitigating safety risks, such as falling objects, collisions, or unexpected robot movements. 
We plan to enhance the control algorithms to allow robots to avoid each other and other obstacles on the wall, preventing such incidents. 
Implementing additional safety features and refining the current system will help ensure user safety while interacting with \textit{corobos}.

\subsection{Durability}
To enhance the practicality and longevity of \textit{corobos}, it is crucial to improve the durability of the robots and their components. 
Due to the significant shocks experienced during transitions, we have observed that thin parts of the attachment tend to break after repeated use. 
We used standard clear resin to fabricate the attachments, but more durable materials would be required to increase their longevity. 
Future research will explore materials and design modifications to enhance wear and tear resistance, contributing to \textit{corobos}' overall performance and reliability in various application scenarios.

\subsection{User Study}
To better understand user experiences and preferences, future work will include conducting user evaluations and usability studies. 
We plan to organize a workshop to gather application ideas and evaluate user perceptions. 
These evaluations will provide valuable insights into the system's effectiveness, user-friendliness, and areas for potential improvement. 
By incorporating user feedback, we aim to refine \textit{corobos} and tailor its features to better meet user needs and expectations.

\section{Conclusion}

We introduced \textit{corobos}, a novel robotic system designed to enable seamless transitions between tabletop and wall surfaces, with the collaborative behavior of mobile robots.
By utilizing a combination of magnetic and mechanical attachments, \textit{corobos} facilitates effective and reliable surface transitions, enhancing dynamic interactions between users and robots.
We presented the underlying concepts, the design process, and the system implementation, including the development of custom robot attachments.
Additionally, we demonstrated various example applications that showcase the potential of \textit{corobos} in different domains, such as workspace organization, education, and spatial display.
Through continued development and refinement, we believe \textit{corobos} has the potential to significantly enhance Human-Robot Interaction by expanding spatial usages and creating more engaging experiences.

\begin{acks}
This work was supported by JSPS KAKENHI Grant Number 21K17778 and Nakayama Future Factory.
\end{acks}

\bibliographystyle{ACM-Reference-Format}
\bibliography{references}

\appendix

\end{document}